\documentclass[11pt]{article}

\usepackage[]{acl}

\usepackage{times}
\usepackage[export]{adjustbox}
\usepackage{latexsym}
\usepackage{colortbl}
\usepackage{color}
\usepackage{caption}
\usepackage{xspace}
\usepackage[T1]{fontenc}

\usepackage[utf8]{inputenc}
\usepackage{graphicx}
\usepackage{subcaption}
\usepackage{adjustbox}
\usepackage{booktabs}
\usepackage{amsmath}
\usepackage{mathtools}
\usepackage[T1]{fontenc}
\usepackage{rotating}
\usepackage{enumitem}
\usepackage{pifont}
\newcommand{\cmark}{\ding{51}}%
\newcommand{\xmark}{\ding{55}}%

\usepackage{amssymb}
\usepackage{multirow}
\usepackage{tabularx}

\usepackage{microtype}

\author{Adnen Abdessaied\thanks{*Corresponding Author.},\, Mihai Bâce,\, Andreas Bulling\\
  Institute for Visualization and Interactive Systems (VIS) \\
  University of Stuttgart, Germany \\
  \texttt{adnen.abdessaied@vis.uni-stuttgart.de} \\
  \texttt{mihai.bace@vis.uni-stuttgart.de} \\
  \texttt{andreas.bulling@vis.uni-stuttgart.de}
}

\usepackage[draft]{changes}
\definecolor{Gray}{gray}{0.9}

\title{Neuro-Symbolic Visual Dialog}
\newcommand{\methodName}{Neuro-Symbolic Visual Dialog (NSVD)\xspace}
\newcommand{\methodNameShort}{NSVD\xspace}

\newcommand{\ffrLong}{First Failure Round (FFR)\xspace}

\begin{document}
\pdfoutput=1
    \maketitle
    \begin{abstract}
We propose \methodName\footnote{Project page: \url{https://perceptualui.org/publications/abdessaied22_coling/}} \textemdash the first method to combine deep learning and symbolic program execution for multi-round visually-grounded reasoning.
\methodNameShort significantly outperforms existing purely-connectionist methods on two key challenges inherent to visual dialog: long-distance co-reference resolution as well as vanishing question-answering performance.
We demonstrate the latter by proposing a more realistic and stricter evaluation scheme in which
we use \textit{predicted} answers for the full dialog history when calculating accuracy.
We describe two variants of our model and show that using this new scheme, our best model achieves an accuracy of $99.72\%$ on CLEVR-Dialog \textemdash a relative improvement of more than $10\%$ over the state of the art \textemdash while only requiring a fraction of training data.
Moreover, we demonstrate that our neuro-symbolic models have a higher mean first failure round, are more robust against incomplete dialog histories, and generalise better not only to dialogs that are up to three times longer than those seen during training but also to unseen question types and scenes.

\end{abstract}
    
    \section{Introduction}

Modelled after human-human communication, visual dialog involves reasoning about a visual scene through multiple question-answering rounds in natural language \cite{Das2019}.
Its multi-round nature gives rise to one of its unresolved key challenges: co-reference resolution \cite{kottur2018visual,Das2019}.
That is, as dialogs unfold over time, questions tend to include more and more pronouns, such as ``it'', ``that'', and ``those''
that have to be resolved to the appropriate previously-mentioned entities in the scene.
Co-reference resolution is profoundly challenging~\cite{Das2019,hu2017learning}, even for models specifically designed for this task \cite{kottur2018visual}.
Existing models follow a purely connectionist approach
and suffer from several limitations:
first, they require large amounts of training data, which is prohibitive for most settings.
Second, these models are not explainable, making it difficult to troubleshoot their logic when co-references are incorrectly resolved.
Finally, current models lack generalisability, in particular for real-world dialogs that include incomplete or inaccurate dialog histories, longer dialogs than those seen during training, or unseen question types.
While neuro-symbolic hybrid models have proven effective as a more robust, explainable, and data-efficient alternative,
e.g. for VQA \cite{nsvqa}, video QA \cite{CLEVRER2020ICLR}, or commonsense reasoning \cite{abs-2006-10022},
they have not yet been explored for visual dialog.

We fill this gap by proposing \methodName \textemdash the first neuro-symbolic method geared towards visual dialog.
Our method combines three novel contributions
to disentangle vision and language understanding from reasoning:
First, it introduces two different program generators: a caption and a question program generator, the former of which induces a program from the caption to initialise the knowledge base of the executor at the beginning of each dialog. 
Second, a question program generator that predicts a program in each round using not only the current question but also the dialog history. 
We describe two variants of this generator: one that uses a question encoder to 
concatenate the caption and question-answer pairs of previous rounds to encode the dialog history, as well as one that stacks them.
Third, a symbolic executor with a dynamic knowledge base keeps track of all entities mentioned in the dialog. 

\methodNameShort~also addresses another limitation of existing models that was ``hidden'' by the dominant evaluation scheme used so far: vanishing question-answering performance over the course of the dialog.
Adopted from VQA \cite{antol2015vqa}, the dominant scheme assumes that the model has full access to the dialog history, in particular all \textit{ground truth} answers.
We argue that this assumption is overly optimistic and overestimates real-world performance on the visual dialog task.
We instead propose a more realistic and stricter evaluation scheme in which prediction in the current round is conditioned on previous \textit{predicted} answers.
This scheme better represents real-world dialogs in which communication partners rarely know whether their previous answers were correct or not.

Through extensive experiments on CLEVR-Dialog \cite{Kottur2019}, we show that our models are significantly better at resolving co-references and at maintaining performance over many rounds.
Using our stricter evaluation scheme, we still achieve an accuracy of $99.72\%$ while requiring only a fraction of the training data. 
Our results further suggest that \methodNameShort
has a higher mean First Failure Round, is more  robust to incomplete dialog histories, and generalises better to dialogs that are up to three times longer than those seen during training as well as to unseen question types and scenes.
The contributions of our work are threefold:
(1) We introduce the first neuro-symbolic visual dialog model that is more robust again incomplete histories, is significantly better at resolving co-references and at maintaining performance over more rounds on CLEVR-Dialog. 
(2) We contribute a new Domain Specific Language (DSL) for CLEVR-Dialog that we augment with ground truth caption and question programs.
(3) We unveil a fundamental limitation of the dominant evaluation scheme for visual dialog models and propose a more realistic and stricter alternative that better represents real-world dialogs.
    
    \section{Related Work}
\paragraph{Neuro-Symbolic Models and Reasoning.}
Many works have used end-to-end connectionist models on the CLEVR dataset \cite{Johnson2017} with varying degrees of success \cite{johnson2017inferring,hu2017learning,perez2018film,hudson2018compositional}.
NS-VQA~\cite{nsvqa} was one of the first neuro-symbolic models on CLEVR, achieving a near-perfect test accuracy.
\citeauthor{Mao2019} proposed NS-CL, a neuro-symbolic VQA network that, in contrast to NS-VQA,
learned simply by looking at images and reading question-answer pairs.
In parallel, other works explored neuro-symbolic models for mono-modal conversational settings \cite{Williams2017,atis,abs-2006-10022}. 
Recently, \citeauthor{SMDataflow2020} introduced a method that represents dialog states as a dataflow graph to better deal with co-references.
Although a number of works have demonstrated the significant potential of neuro-symbolic methods for mono-modal task-oriented dialog settings or tasks at the intersection of computer vision and natural language processing, we are the first to use them for the multi-modal visual dialog task.

\paragraph{Visual Dialog.}
\citeauthor{Das2019} introduced early visual dialog models that used an encoder-decoder approach to rank a set of possible answers. Others explored explicit reasoning based on the dialog structure \cite{zheng2019reasoning,niu2019recursive,gan-etal-2019-multi}. 
However, these models focused mainly on the real-world VisDial dataset \cite{Das2019}. Although popular, this dataset is not well-suited to study a key challenge of visual dialog, i.e. co-reference resolution, because it lacks complete annotation of all images and dialogs.
Several works have focused on co-reference resolution in videos \cite{ramanathan2014linking,rohrbach2017generating} and 3D data \cite{kong2014you}. 
\citeauthor{Kottur2019} introduced CLEVR-Dialog -- a fully-annotated diagnostic dataset for multi-round visual reasoning with a grammar grounded in the CLEVR scene graphs.
More recently, \citeauthor{shah-etal-2020-reasoning} introduced models that build on the  Memory, Attention, and Composition (MAC) network \cite{hudson2018compositional}.
Although their models achieved promising results on CLEVR-Dialog, they are computationally and memory inefficient \cite{shah-etal-2020-reasoning}.
While the neuro-symbolic approach has significant potential to address these shortcomings, it has not yet been explored for visual dialog.

    \section{Method}
\begin{figure*}[!t]
        \centering
        \scalebox{1}[1]{
        \includegraphics[width=\textwidth]{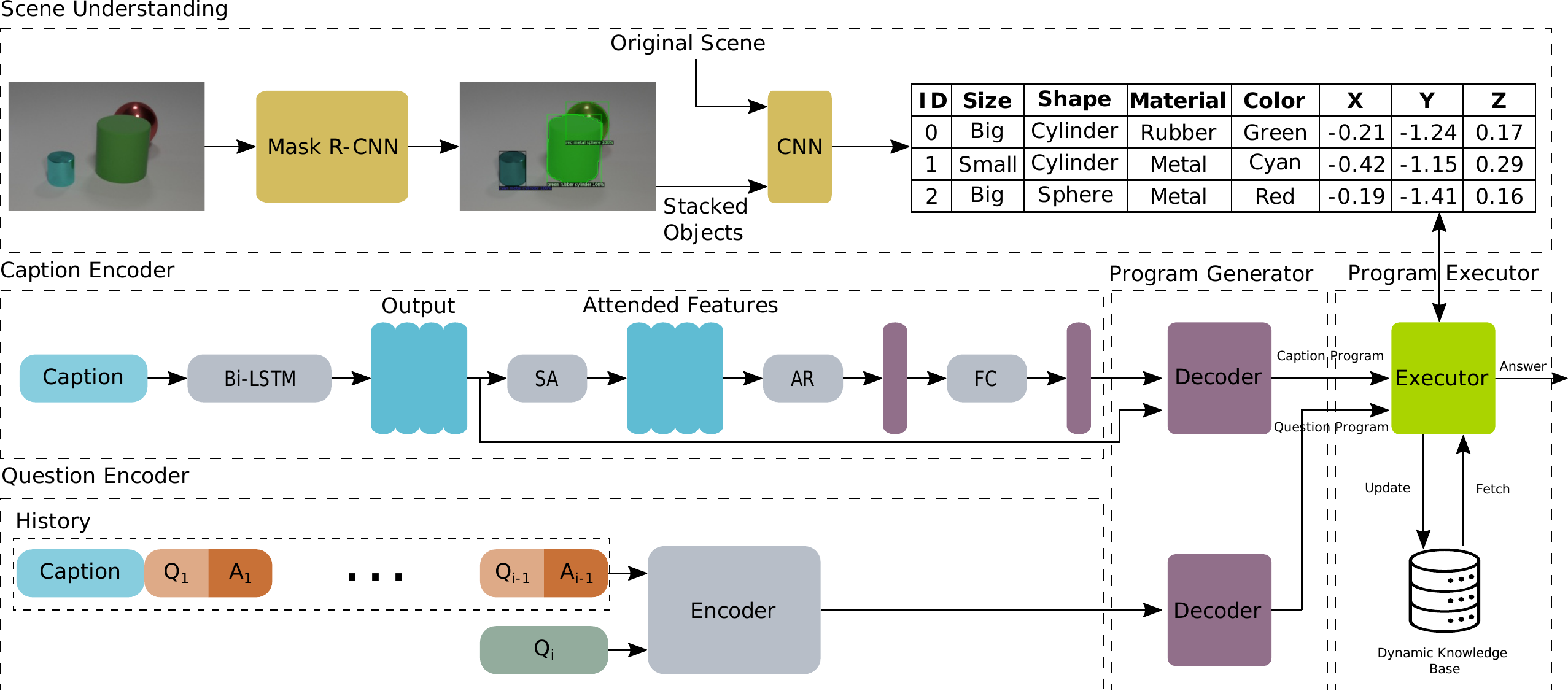}
        }
        \caption{Overview of our method: first, a structured scene representation is created. Then, a caption program is induced and run by our executor to initialise its knowledge base. At each subsequent round, the question and the history are used to induce a program that answers the question and updates the dynamic knowledge base.}
        \label{fig:overview}
\end{figure*}

Our method consists of four components (see \autoref{fig:overview}): a scene understanding method, a program generator with caption and question encoders and a decoder, and a symbolic program executor with a dynamic knowledge base.

\begin{figure*}[!t]
        \centering
        \scalebox{1}[1]{

        \includegraphics[width=\textwidth]{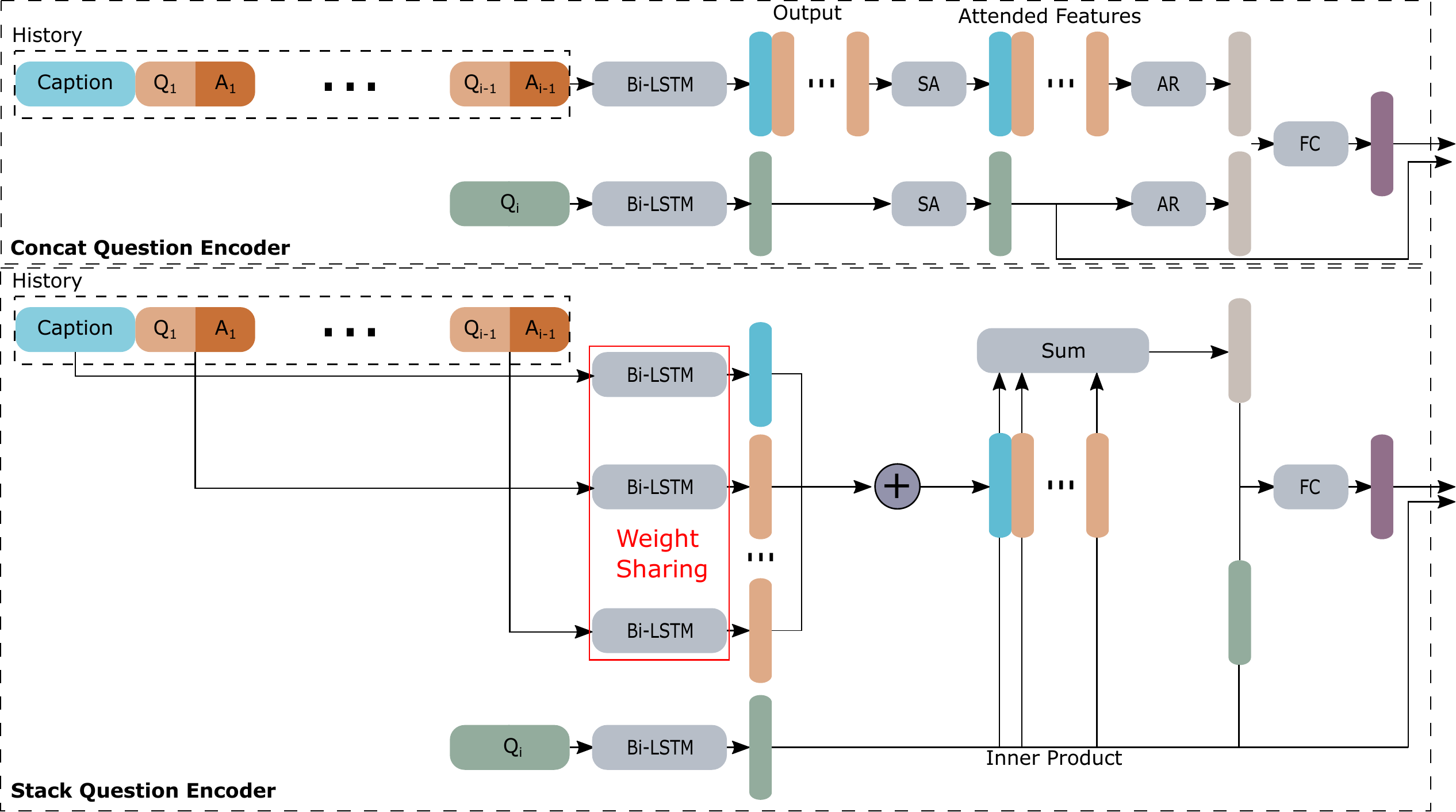}
        }
        \caption{
        \textbf{Top:} The concatenative encoder (``concat'') takes the question and the concatenated previous rounds as input and outputs a latent vector and a question representation to the decoder. \textbf{Bottom:} The stacking encoder (``stack'') takes the question and the previous rounds as input and attends to each round separately. Then, it outputs a latent vector and a question representation to the decoder.} 
        \label{fig:method}
\end{figure*}

\paragraph{Scene Understanding.}

We used a pre-trained Mask R-CNN \cite{he2017mask} to predict segmentation masks and attributes (colour, shape, material, size) for each entity in the visual scene.
We then learned the 3D coordinates of each segment paired with the original image using a ResNet-34 \cite{he2016deep}.
The Mask R-CNN was pre-trained on the CLEVR-mini dataset \cite{nsvqa}.

\paragraph{DSL for CLEVR-Dialog.}
Because CLEVR-Dialog implements its own grammar and vocabulary, we designed a novel domain-specific language (DSL) for it by implementing a collection of deterministic functions in Python that our symbolic executor can run over a CLEVR scene.
In previous works \cite{johnson2017inferring,nsvqa,Mao2019}, these functional modules shared the same input/output interface and were arranged one after another to predict the answer.
Instead, we followed a stricter approach by executing only one function that expects a \textit{different} number of input arguments to answer a particular question.
The full list of our functions, their arguments, and expected output can be found in Appendix \ref{sec:appendix_dsl}.

\paragraph{Program Generation.}
\begin{figure}[!t]
    \begin{minipage}{1\linewidth}
        \centering
        \scalebox{1}[1]{
        \includegraphics[width=\textwidth]{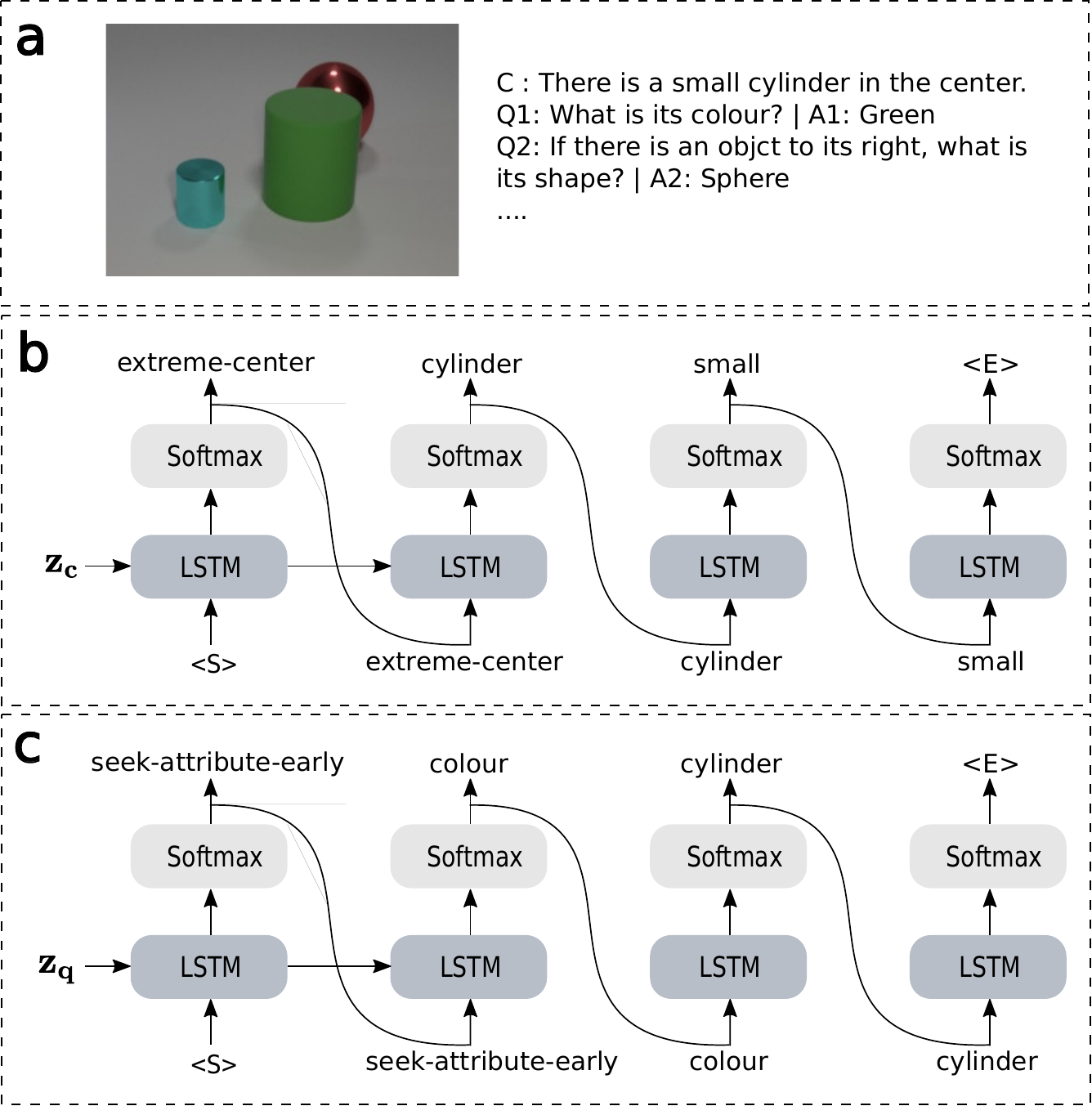}
        }
        \caption{\textbf{a:} An example of a CLEVR-Dialog instance with a caption and two rounds. \textbf{b:} Inference sample of the caption program generation. \textbf{c:} Inference sample of the question program generation on the first round.}
        \label{fig:decoder}
    \end{minipage}
\end{figure}

Semantic parsing methods were shown to be effective for mapping sentences to logical forms via a knowledge base or a program \cite{guu-etal-2017-language,liang-etal-2013-learning,atis}.
We adopted this approach and used a sequence-to-sequence model with an encoder-decoder structure to generate the programs.
Although we used the same decoder, we propose two types of encoders that differ in the way they encode the dialog history (\autoref{fig:overview}).

\textit{Caption Encoder.}
The caption encoder first embeds the caption tokens into a 300-dim. space to give $\{\mathbf{w_{c_j}}\}_{j=1}^{n_c}$ that are then fed into a bi-directional LSTM \cite{lstm}. 
The self-attended \cite{transformers} LSTM outputs $\mathbf{C} = [\mathbf{c_1}, ..., \mathbf{c_{n_c}}]$ are reduced following:
\begin{align*}
    \mathbf{a} &= \mathrm{softmax}(\textrm{MLP}(\mathbf{C})) \\
    \mathbf{\bar{z}_c} &= \sum_{i=1}^{n_c} a_i \mathbf{c_i}.
\end{align*}
Finally, the latent vector $\mathbf{z_c} = \textrm{Linear}(\mathbf{\bar{z}_c})$ and the LSTM output $\mathbf{C}$ are passed to the decoder.

\textit{Question Encoders.}
To generate the question program at the current round $i$, we use not only the question $Q_i$ but also the history $H_i = [C, Q_1, A_1, .., Q_{i-1}, A_{i-1}]$ of previous question-answer pairs including the caption.
We propose two different encoders based on how the question interacts with the history.

-- Concat Encoder: The concat encoder is similar in structure to the caption encoder.
The caption and the question-answer pairs of previous rounds are concatenated to form the history $H_i$.
Then, the tokens of the current question $Q_i$ and history $H_i$ are embedded into a $300$-dim. space to give  $\{\mathbf{w_{q_j}^{(i)}}\}_{j=1}^{n_q}$ and $\{\mathbf{w_{h_j}^{(i)}}\}_{j=1}^{n_h}$, respectively, which are then processed by two separate bi-directional LSTMs (\autoref{fig:method}).
The reduced attended question and history features $\mathbf{\bar{z}_q}$ and $\mathbf{\bar{z}_h}$ are obtained in a similar manner to $\mathbf{\bar{z}_c}$.
Finally, $\mathbf{\bar{z}_q}$ and $\mathbf{\bar{z}_h}$ are concatenated and linearly transformed to produce the question latent vector $\mathbf{z_q} = \mathrm{Linear}([\mathbf{\bar{z}_q}, \mathbf{\bar{z}_h}]).$
Similarly, $\mathbf{z_q}$ and the question LSTM output $\mathbf{Q_i} = [\mathbf{q_1}, ..., \mathbf{q_{n_q}}]$ are passed to the decoder.

-- Stack Encoder: The approach of concatenating the question-answer pairs to form the history suffers from two main drawbacks.
First, since the history is processed by an LSTM, its encoding becomes inefficient, in particular for later rounds as the LSTM tends to forget crucial information that was mentioned in the first rounds. Second, this approach does not scale well for longer dialogs as it becomes computationally and memory demanding 
especially because we use self-attention \cite{transformers} at a later stage in the concat encoder.
To overcome these limitations, we introduce the stack encoder that separately encodes each question-answer pair in order to equally preserve the information from all previous rounds. 
The question $Q_i$ and each previous round $R_{j<i}$, including the caption, are embedded 
then processed by separate bidirectional LSTMs (\autoref{fig:method}).
The last hidden states are used as feature representations of the question and previous rounds, i.e. 
\begin{equation*}
    \mathbf{q} = [\overrightarrow{\mathbf{h}_{Q_i}}, \overleftarrow{\mathbf{h}_{Q_i}}] \textrm{ and } \mathbf{r_{j<i}} = [\overrightarrow{\mathbf{h}_{R_{j<i}}}, \overleftarrow{\mathbf{h}_{R_{j<i}}}],
\end{equation*}

where $\overrightarrow{\mathbf{h_{(.)}}}$ and  $\overleftarrow{\mathbf{h_{(.)}}}$ are the bi-directional LSTM's last forward and backward hidden states, respectively.  
$\mathbf{\bar{z}_h}$ is obtained by applying an inner-product attention between the question and history features:
\begin{align*}
     \mathbf{a} &= \mathrm{softmax}(\mathbf{{q}^T} \mathbf{H}),\\
     \mathbf{H} &= [\mathbf{r_{0}}, ..., \mathbf{r_{i-1}}].
\end{align*}
   
 Finally, $\mathbf{q}$ and $\mathbf{\bar{z}_h} = \sum_{j=1}^{i-1} a_j \mathbf{r_j}$ are concatenated and linearly transformed to produce the latent question vector $\mathbf{z_q} = \mathrm{Linear}([\mathbf{q}, \mathbf{\bar{z}_h}]).$
Similarly, $\mathbf{z_q}$ and the question LSTM output $\mathbf{Q} = [\mathbf{q_1}, ..., \mathbf{q_{n_q}}]$ are passed to the decoder.

\textit{Decoder.}
We use the same decoder architecture to generate all the caption as well as the question programs.
First, the ground truth program sequence $Y_{i}$ of the $i$-th dialog round is embedded into a $300$-dim. space to give $\{\mathbf{w_{y_j}^{(i)}}\}_{j=1}^{n_y}$ which are then processed by a simple LSTM whose hidden states are initialised by the encoder latent vector, i.e.  $\mathbf{z_c}$ or $\mathbf{z_q}$. 
The output $\mathbf{P}$ of the LSTM is used with the encoder output, i.e. $\mathbf{C}$ or $\mathbf{Q}$, to generate a context vector $\mathbf{\Delta}$ following:
\begin{align*}
    \mathbf{A} &= \mathrm{softmax}(\mathbf{Q^T P}),\\
    \mathbf{\Delta} &= \mathbf{A^T Q}.
\end{align*}
Finally, the context vector $\mathbf{\Delta}$ is concatenated with the program output and the result is mapped to the program vocabulary dimension followed by a softmax function to obtain a distribution for the current program token $y_j$, i.e. 
\begin{equation*}
    \begin{split}
        p(y_j|Y_{[1:j-1]}; Q_i, H_i,) \sim \mathrm{softmax}(& \\ \mathrm{Linear}(\mathrm{tanh}([\mathbf{P}, \mathbf{\Delta}]))),
    \end{split}
\end{equation*}
where $Y_{[1:j-1]}$ is the sequence of previous ground truth program tokens.
For training, we follow the teacher forcing strategy by \citeauthor{williams1989learning}. For inference, we first start with the \texttt{<S>} and sequentially generate the next program token until we reach the end token \texttt{<E>}. \autoref{fig:decoder} illustrates an example of the CLEVR-Dialog dataset alongside the generated programs, i.e. the program \texttt{extreme-centre(cylinder, small)} was generated from the caption ``\textit{There is a small cylinder in the center}'' to initialize our executor and its knowledge base and the program \texttt{seek-attribute-early(colour, cylinder)} was generated to answer the first question of the dialog ``\textit{What is its colour?}''. Our full DSL grammar and further concrete examples can be found in the Appendices \ref{sec:appendix_dsl} and \ref{sec:qualitative_ana}, respectively.

\paragraph{Executor.}

We add a dynamic knowledge base to the symbolic executor to keep track of the previously-mentioned entities in the dialog. 
It is initialised at the beginning of each dialog by executing the caption program. 
For instance, by executing the caption program \texttt{extreme-centre(cylinder, small)}, the executor searches for the centre entity satisfying the function's arguments and stores it in the knowledge base under the handle \texttt{small-cylinder}. 
The executor interacts with its knowledge base via two main operations:

\paragraph{\texttt{fetch}.}
The \texttt{fetch}-operation is performed when executing a function that requires co-reference resolution. 
Given a set of attributes, the executor fetches the appropriate entity in the knowledge base by searching the stored handles. 
For example \texttt{seek-attribute-early(colour, cylinder)} first searches the previously stored handles and fetches the corresponding entity (in our example that is the cylinder mentioned in the caption with the handle \texttt{small-cylinder}) and then queries its colour to answer the question.

\paragraph{\texttt{update}.}
The \texttt{update}-operation is performed after each question function. We differentiate between four update types:
\begin{enumerate}[wide,labelindent=0cm,leftmargin=0cm]
    \itemsep0em 
    \item \textit{Handle update:} If a fetched entity is referenced by a new attribute, its handle in the knowledge base should be updated accordingly. If the colour of the previous cylinder is red, then its handle changes from \texttt{small-cylinder} to \texttt{small-cylinder-red}.
    \item \textit{Conversation subject update:} If the question program addresses a new entity, the latter becomes the new conversation subject. In our example, the conversation subject is still the small red cylinder. However, the question program \texttt{exist-obj-exclude-early(colour, small, cylinder)} searches for other potential entities that share the same colour as the previous small cylinder. If there is one, it becomes the new conversation subject.
    \item \textit{Seen entities update:} Each time a new entity is addressed, the executor saves it in its knowledge base together with the appropriate handle.
    \item \textit{Groups update:} Some questions refer to a group of entities, e.g. \texttt{count-attribute(red)} counts all red entities in the scene. These sets might be relevant for subsequent questions, e.g. \texttt{count-attribute-group(large)} counts how many of the previous red entities are large.
\end{enumerate}
    
    \section{Experiments}
We modified the publicly available code for CLEVR-Dialog \cite{Kottur2019}
to generate datasets with ground truth caption and question programs required to train our program generators.
Similar to \cite{Kottur2019}, we used the $70,000$ training and $15,000$ validation CLEVR images as our visual groundings when generating the dialogs. 
We left out the CLEVR test images because they lack ground truth scene annotations.
For each image, we generated five dialogs each consisting of $L = 10$ question-answer rounds as in \cite{Kottur2019}. 
We used $1,000$ training images and their corresponding dialogs to create a validation set and tested our models and the baselines on the dialogs generated using the CLEVR validation images.

\paragraph{Performance Evaluation.}
Alongside the answer accuracy, the \ffrLong, i.e. the number of dialog rounds necessary for a model to make its first mistake, is commonly used to evaluate visual dialog models. Although popular, this metric has one major limitation as it only allows us to compare the performance of models across datasets with the same dialog length but not across datasets with different ones. Thus, we propose the \textit{Normalised First Failure Round} $\textrm{(NFFR)} \in [0, 1]$ as an improvement and use it alongside the answer accuracy to assess the performance of all models. See Appendix \ref{sec:appendix_ffr} for more details.

\paragraph{History during Evaluation.}

One key limitation in the way that visual dialog models are currently evaluated is the use of \textit{ground truth} answers when calculating the correctness of an answer in any given round \cite{kottur2018visual, Das2019, shah-etal-2020-reasoning}. 
The problem of this approach is that it leads to overly-optimistic performance that do not reflect the true capabilities of the models in real-world scenarios:
in real-world dialogs, full information on which previous answers were correct or not is typically not available.
We instead propose to condition the generation of the current answer on all previous \textit{predicted} answers.
This new evaluation scheme is geared to the visual dialog task, better represents real-world use, and is stricter.
We call these evaluation schemes ``Hist. + GT'' and ``Hist. + Pred.'', respectively.

    \section{Results}

\begin{table}
\centering
\scalebox{0.75}[0.75]{

\begin{tabular}{l||cc||cc}
\hline
\multirow{2}*{\textbf{Model}}& \multicolumn{2}{c||}{\textbf{Hist. + GT}} & \multicolumn{2}{c}{\textbf{Hist. + Pred.}}\\
\cmidrule(r){2-3} \cmidrule(r){4-5}

& \textbf{Acc.}&  $\mathbf{{NFFR} \uparrow}$& \textbf{Acc.}&  $\mathbf{{NFFR} \uparrow}$\\
\hline
MAC-CQ         & $97.34^\ddag$       & $0.92$\,\,  & $41.10\,\,$      & $0.15$\,\, \\
$\quad$+ CAA   & $97.87^\ddag$       & $0.94$\,\,  & $89.39^\ddag$  & $0.75$\,\, \\
$\quad$+ MTM   & $97.58^\ddag$       & $0.92$\,\,  & $70.39^\ddag$  & $0.46$\,\, \\
\hline
HCN            & $75.88\,\,$         & $0.34$\,\,  & $74.42^\ddag$  & $0.32$\,\, \\
\rowcolor{Gray}
NSVD-concat & $99.59^\ddag$           &$0.98$\,\,           & $99.59^\ddag$          & $0.98$\,\,           \\
\rowcolor{Gray}
NSVD-stack & $\mathbf{99.72}^\ddag$  & $\mathbf{0.99}$\,\, & $\mathbf{99.72}^\ddag$ & $\mathbf{0.99}$\,\,  \\ 
\hline
\end{tabular}
}
\caption{Performance comparison of our models with the state of the art on CLEVR-Dialog \textit{test}.
Results are shown for both ``Hist. + GT'' and ``Hist. + Pred.'' 
Our proposed models are highlighted in grey; best performance is in bold. $\ddag$ represents $p < 0.00001$ compared to the second best score in the respective column.
}
\label{tab:overall_perfomance}
\end{table}

\paragraph{Visual Dialog Performance.}
\label{sec:overall_performance}
After validating our implemented logic (see Appendix \ref{sec:logic_analysis}), we compared the performance of our models with the visual dialog MAC networks introduced by \citet{shah-etal-2020-reasoning}. 
We limited our comparison to their top three performing models given that these outperformed the previous state of the art \cite{kottur2018visual} by $30\%$ in accuracy.
Furthermore, we compared our models to the Hybrid Code Networks (HCN) \cite{Williams2017} that also operate on symbolic dialog state representation but follow a different approach to parse programs than our generative one. They represent programs as templates in an action space and select the one with the highest probability during inference. This action space might become intractable if the DSL has 
many functions and arguments.

\autoref{tab:overall_perfomance} shows the performance of our models
and the baselines on the test split using both evaluation schemes (``Hist. + GT'' and ``Hist. + Pred.''). 
As the table shows, our models achieve new state-of-the-art performance with \textit{NSVD-stack} topping with an overall accuracy of $99.72\%$ and a ${\textrm{NFFR}} = 0.99$.
The high ${\textrm{NFFR}}$ demonstrates our models' ability to answer correctly across all rounds of the dialogs with only few failures in between.
More fine-grained evaluations (e.g. on individual rounds, question categories and types) are available in the Appendix \ref{sec:finegrained_eval}.
While \autoref{tab:overall_perfomance} shows results obtained when training on the entire dataset, our method achieves the same performance when trained on only $20\%$ of the data, while the performance of other methods deteriorate significantly with less data as shown in Appendix \ref{sec:data_efficiency}.
\begin{figure*}[!t]
        \centering
        \scalebox{1.0}[1.0]{
            \includegraphics[width=\textwidth]{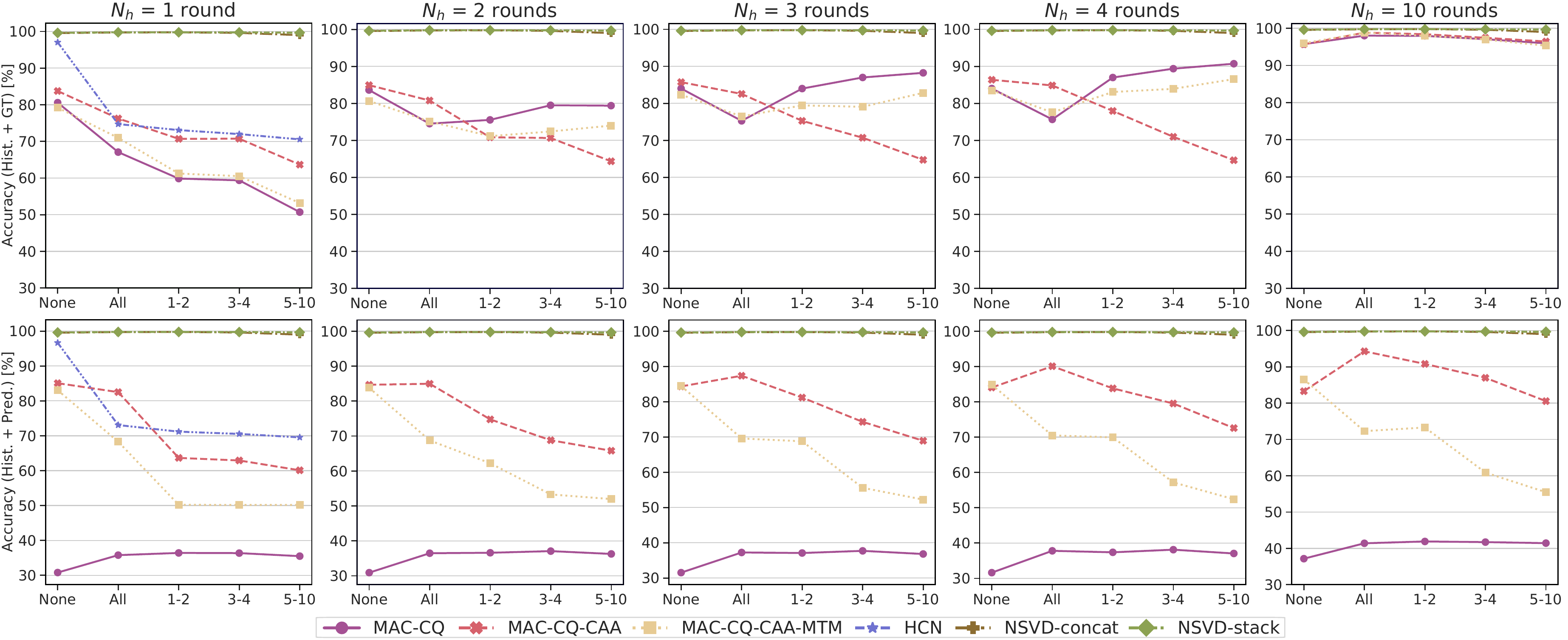}
        }
        \caption{Robustness for different co-reference distance bins and varying number of rounds in the history. All  models were trained with full histories. Independent of the evaluation scheme, the performance of our models is only slightly affected by incomplete histories across all bins. In contrast, performance of the baselines deteriorates when the full history is not available, especially for questions with large co-reference distances.}
        \label{fig:hist_vs_coreference}
\end{figure*}

\paragraph{History Length vs Co-reference Distance.}
CLEVR-Dialog provides co-reference distances for each question, i.e. the number of rounds between the current and earlier mention of an entity in a question. 
A co-reference distance of $1$ means that the co-referent was mentioned in the previous question while a co-reference distance of $10$ means that the question at round $10$ refers to an entity in the caption.
``All'' and ``None'' mean that the question either depends on all previous rounds or is stand-alone, i.e. it does not depend on the history.

To assess performance with respect to the co-reference distance, we evaluated accuracy on different co-reference distance bins.
In our evaluations, we further limited the histories to the last $N_h$ question-answer rounds to assess the robustness of the models to incomplete dialog histories. All of the models were trained with histories containing all previous rounds except for HCN \cite{Williams2017} that only uses the last round.

Our models consistently achieve a performance of over $99\%$ across all co-reference distance bins, independent of the evaluation scheme (\autoref{fig:hist_vs_coreference}).
Furthermore, their performance is only slightly affected by incomplete dialog histories.
Contrarily, performance of all connectionist baselines deteriorates quickly without access to the complete history.
This deterioration is more conspicuous when the ``Hist. + Pred.'' evaluation scheme is used (second row of \autoref{fig:hist_vs_coreference}).
However, their performance is consistent with the difficulty levels of the co-reference distance bins, i.e. the accuracy decreases with increasing co-reference distance. In contrast, this behaviour is not reflected by the popular evaluation approach ``Hist. + GT'' (first row, middle three plots of \autoref{fig:hist_vs_coreference}).
The most likely reason for this is
that, as is currently common practice when evaluating visual dialog models, the ground truth answers of all previous rounds are used for prediction. 

\begin{table*}[!h]
    \begin{minipage}{1\linewidth}
      \centering
        \scalebox{0.63}[0.63]{
            \begin{tabular}{cl||cc|cc||cc|cc||cc|cc}
            \hline
            & \multirow{3}*{\textbf{Model}} & \multicolumn{4}{c||}{$\mathbf{N_{objects} = 10}$} & \multicolumn{4}{c||}{$\mathbf{N_{objects} = 15}$} & \multicolumn{4}{c}{$\mathbf{N_{objects} = 20}$} \\
            \cmidrule(r){3-6} \cmidrule(r){7-10}  \cmidrule(r){11-14}
            
            & & \multicolumn{2}{c|}{\textbf{Without FT}} & \multicolumn{2}{c||}{\textbf{After FT}} & \multicolumn{2}{c|}{\textbf{Without FT}} & \multicolumn{2}{c||}{\textbf{After FT}} &  \multicolumn{2}{c|}{\textbf{Without FT}} & \multicolumn{2}{c}{\textbf{After FT}} \\
            \cmidrule(r){3-4} \cmidrule(r){5-6}  \cmidrule(r){7-8} \cmidrule(r){9-10} \cmidrule(r){11-12}  \cmidrule(r){13-14}

            & & \textbf{Acc.}& $\mathbf{{NFFR} \uparrow}$&  \textbf{Acc.}& $\mathbf{{NFFR} \uparrow}$&  \textbf{Acc.}& $\mathbf{{NFFR} \uparrow}$& \textbf{Acc.}& $\mathbf{{NFFR} \uparrow}$&  \textbf{Acc.}& $\mathbf{{NFFR} \uparrow}$&  \textbf{Acc.}& $\mathbf{{NFFR} \uparrow}$\\
            \hline
            \multirow{6}*{\rotatebox{90}{\textbf{Hist. + GT}}}  
            &\multicolumn{1}{|l||}{MAC-CQ}     & ${38.52^\dag}$ & ${0.14}$\,\,        &  ${53.49^*}$ & $0.21$\,\,        & $\mathbf{36.87^*}$ & $\mathbf{0.13}$\,\,  & $48.89^*$ & $0.19$\,\,        &  $\mathbf{36.12}^*$ & $\mathbf{0.13}$\,\,        & $47.44^*$ & $0.18$\,\, \\
            &\multicolumn{1}{|l||}{$\quad$+ CAA}  & $37.72^\ddag$ & $0.14$\,\,        &  $53.35^\dag$ & $0.21$\,\,        & $36.82^\dag$ & $0.13$\,\, & $48.67^\dag$ & $0.18$\,\,        &  $35.52^\ddag$ & $0.13$\,\,        & $47.43^\dag$ & $0.18$\,\, \\
            &\multicolumn{1}{|l||}{$\quad$+ MTM}& $\mathbf{38.59}^*$ & $\mathbf{0.14}$\,\,         &  $52.62$\,\, & $0.21$\,\,        & $36.22^\ddag$ & $0.13$\,\,& $47.41$\,\, & $0.17$\,\,         &  $36.01^*$ & $0.13$\,\,        & $46.54$\,\, & $0.17$\,\,\\
            \cline{2-14}

            &\multicolumn{1}{|l||}{HCN}     & $19.59$\,\, & $0.11$\,\,        &  $73.07^\ddag$ & $0.30$\,\,         & $14.42$\,\, & $0.11$\,\,   & $56.65^\ddag$ & $0.22$\,\,         &  $12.33$\,\, & $0.11$\,\,         & $53.14^\ddag$ & $0.19$\,\,  \\
            &\multicolumn{1}{|l||}{\cellcolor{gray!25}NSVD-concat}& \cellcolor{gray!25}$25.05^*$           &\cellcolor{gray!25}$0.12$\,\,  & \cellcolor{gray!25}$99.32^\ddag$ & \cellcolor{gray!25}$0.97$\,\, &\cellcolor{gray!25}$18.51^*$ & \cellcolor{gray!25}$0.11\,\,$ & \cellcolor{gray!25}$70.59^\ddag$ & \cellcolor{gray!25}$0.44$\,\,  & \cellcolor{gray!25}$15.67^\ddag$ & \cellcolor{gray!25}$0.11$\,\, &\cellcolor{gray!25}$64.82^\ddag$ & \cellcolor{gray!25}$0.38\,\,$ \\
            &\multicolumn{1}{|l||}{\cellcolor{gray!25}NSVD-stack} &  \cellcolor{gray!25}${24.95^\ddag}$ & \cellcolor{gray!25}${0.12}$\,\, & \cellcolor{gray!25}$\mathbf{99.33^*}$ & \cellcolor{gray!25}$\mathbf{0.98}$\,\, & \cellcolor{gray!25} $18.45^\ddag$\,\, & \cellcolor{gray!25} ${0.11}$\,\,\,\, &  \cellcolor{gray!25}$\mathbf{70.62^*}$ & \cellcolor{gray!25}$\mathbf{0.44}$\,\, & \cellcolor{gray!25}${15.65^*}$ & \cellcolor{gray!25}${0.11}$\,\, & \cellcolor{gray!25} $\mathbf{64.95^*}$\,\, & \cellcolor{gray!25} $\mathbf{0.38}$\,\,\,\\ 
            \hline
            \multirow{6}*{\rotatebox{90}{\textbf{Hist. + Pred.}}}  
            
            &\multicolumn{1}{|l||}{MAC-CQ}     & $\mathbf{37.74^*}$ & $\mathbf{0.14}$\,\,        &  $52.26^\dag$ & $0.21$\,\,        & $36.32^*$ & $0.12$\,\,  & $47.58^*$ & $0.18$\,\,        &  $35.36^\ddag$ & $0.13$\,\,        & $46.30^\dag$ & $0.18$\,\, \\
            &\multicolumn{1}{|l||}{$\quad$+ CAA}  & $37.70^\ddag$ & $0.13$\,\,        &  $51.36^\dag$ & $0.21$\,\,        & $\mathbf{36.76^*}$ & $0.13$\,\, & $47.08^\ddag$ & $0.18$\,\,        &  $\mathbf{35.57^*}$ & $\mathbf{0.13}$\,\,        & $45.56^\ddag$ & $0.17$\,\, \\
            &\multicolumn{1}{|l||}{$\quad$+ MTM}& $36.29^\ddag$ & $0.14$\,\,         &  $50.58$\,\, & $0.20$\,\,        & $35.62^\ddag$ & $\mathbf{0.14}$\,\,& $45.67$\,\, & $0.17$\,\,         &  $33.98^\ddag$ & $0.13$\,\,        & $44.02$\,\, & $0.17$\,\,\\
            \cline{2-14}

            &\multicolumn{1}{|l||}{HCN}     & $19.50$\,\, & $0.11$\,\,        &  $71.55^\ddag$ & $0.29$\,\,         & $14.40$\,\, & $0.11$\,\,   & $55.55^\ddag$ & $0.21$\,\,         &  $12.26$\,\, & $0.11$\,\,         & $51.95^\ddag$ & $0.19$\,\,  \\
            &\multicolumn{1}{|l||}{\cellcolor{gray!25}NSVD-concat}& \cellcolor{gray!25}$25.05^*$           &\cellcolor{gray!25}$0.12$\,\,  & \cellcolor{gray!25}$99.32^\ddag$ & \cellcolor{gray!25}$0.97$\,\ &\cellcolor{gray!25}$18.51^*$ & \cellcolor{gray!25}$0.11\,\,$ & \cellcolor{gray!25}$70.59^\ddag$ & \cellcolor{gray!25}$0.44$\,\,  & \cellcolor{gray!25}$15.67^\ddag$ & \cellcolor{gray!25}$0.11$\,\, &\cellcolor{gray!25}$64.82^\ddag$ & \cellcolor{gray!25}$0.38\,\,$ \\
            &\multicolumn{1}{|l||}{\cellcolor{gray!25}NSVD-stack} &  \cellcolor{gray!25}${24.95^\ddag}$ & \cellcolor{gray!25}${0.12}$\,\, & \cellcolor{gray!25}$\mathbf{99.33^*}$ & \cellcolor{gray!25}$\mathbf{0.98}$\,\, & \cellcolor{gray!25} ${18.45^\ddag}$\,\, & \cellcolor{gray!25} ${0.11}$\,\,\,\, &  \cellcolor{gray!25}$\mathbf{70.62^*}$ & \cellcolor{gray!25}$\mathbf{0.44}$\,\, & \cellcolor{gray!25}${15.65^*}$ & \cellcolor{gray!25}${0.11}$\,\, & \cellcolor{gray!25} {$\mathbf{64.95^*}$} & \cellcolor{gray!25} $\mathbf{0.38}\,\,\,$\\ 
            \hline
            \end{tabular}
        }
        \caption{Results when training on simple scenes and testing on more complex ones. Best results in bold.
        $\ddag, \dag$, and $*$ represent $p < 0.00001$, $p < 0.05$ and $p \geq 0.05$ compared to the second best score in each column, respectively.}
        \label{tab:unseen_scenes}
    \end{minipage} 
\end{table*}

\begin{figure*}[ht]
    \centering
    \scalebox{0.8}[0.8]{
    \includegraphics[width=0.25\linewidth]{
        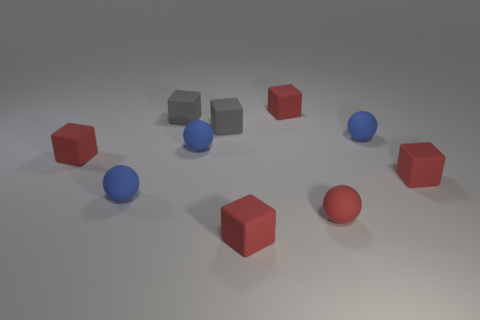}
    }\hfil
    \scalebox{0.8}[0.8]{

    \includegraphics[width=0.25\linewidth]{
        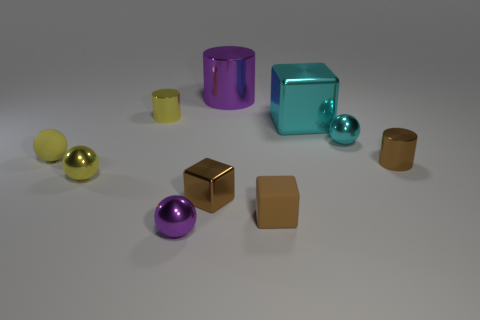}
    }\hfil
    \scalebox{0.8}[0.8]{

    \includegraphics[width=0.25\linewidth]{
        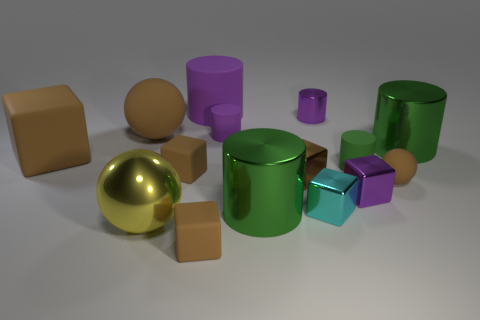}
    }\hfil
    \scalebox{0.8}[0.8]{

    \includegraphics[width=0.25\linewidth]{
        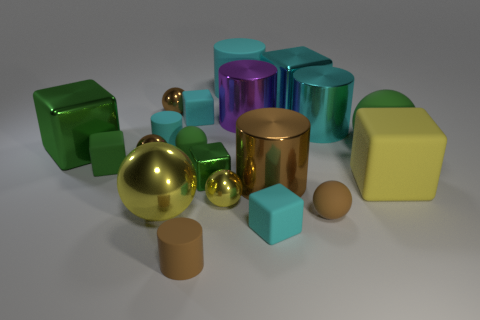}
    }\hfil
    \caption{\textbf{From left to right:} Samples of a training image, fine-tuning image with 10 objects, fine-tuning image with 15 objects, and fine-tuning image with 20 objects.}
\label{fig:many_objects_examples}
\end{figure*}
\paragraph{Generalisation to Unseen Scenes and Attributes.}
In previous experiments, our training and validation sets had similar distributions both in the number of objects (between three and $10$) as well their sizes, shapes, colours, and materials.
To further test generalisability, we created a new training set consisting of $1500$ images in which we restricted the type of objects to small, rubber cubes and spheres with the colours grey, red, or blue. We kept the number of objects in this dataset between three and $10$.
For testing, we generated three datasets consisting of $1000$ images each in which we allowed all CLEVR object classes (cubes, spheres and cylinders) and materials (rubber, metal) to appear.
However, we excluded the training colours and increased the number of objects $N_{objects}$ in each one to $10$, $15$, and $20$, respectively. 
Finally, we generated three fine-tuning datasets containing $1500$ images each in a similar way to the testing ones.
\autoref{fig:many_objects_examples} illustrates some examples of our new images.
As in \cite{Kottur2019}, all dialogs had a length of $10$ rounds.

We can see that the purely-connectionist models outperform the neuro-symbolic ones without fine-tuning in all scene complexities (\autoref{tab:unseen_scenes}). 
This outcome is expected since these models rely on a Mask-RCNN to understand the scenes. By increasing their complexities, i.e. more objects and attributes, the Mask-RCNN fails to accurately reconstruct these scenes which is reflected by the poor test accuracies.
However, after fine-tuning, neuro-symbolic models are the best performing with our best model \textit{NSVD-stack} scoring $99.33\%$, $70.62\%$, and $64.95\%$ on the test datasets with $10$, $15$, and $20$ objects, respectively.

\begin{figure}[!t]
    \begin{minipage}{1\linewidth}
        \centering
        \scalebox{1}[1]{
            \includegraphics[width=\textwidth]{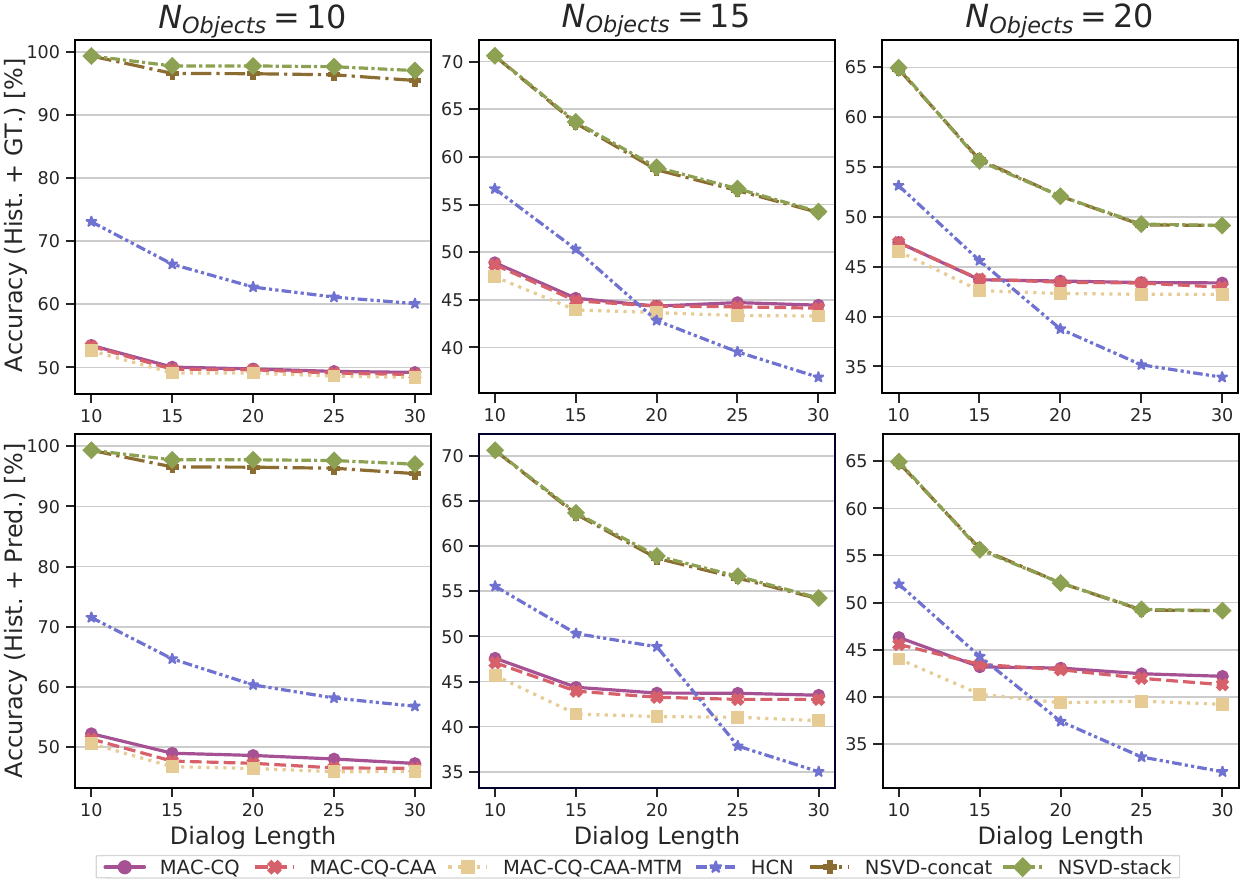}
        }
        \caption{Answer accuracy for different dialog lengths and scene complexities.
        Our models generalise better to longer dialogs \textit{without} the need for any fine-tuning.
        }
        \label{fig:longer_dialogs}
    \end{minipage}
\end{figure}

\paragraph{Generalisation to Longer Dialogs.}
Typically, evaluation of visual dialog models is limited to dialogs having the same length as the training data, i.e. $L = 10$ \cite{kottur2018visual,Das2019,shah-etal-2020-reasoning}.
In order to assess the generalisation capabilities of our models to longer dialogs, we used the testing images of the previous experiment to generate three dialog datasets with increasing numbers of rounds, i.e. $L = 15$, $20$, $25$, and $30$, respectively. That is, our testing datasets for this experiment not only contain dialogs that are up to three times the length of the training dialogs but also visual scenarios never seen during training.
Finally, we evaluated the best fine-tuned models of the previous experiment on this data \textit{without} fine-tuning them again on longer dialogs.
\autoref{fig:longer_dialogs} shows that our models generalise better across all datasets for both evaluation schemes.
As expected, the performance of all models decreases with longer dialogs and more complex scenes.
However, our models suffer less and still significantly outperform all baselines in all test scenarios.
Our best model \textit{NSVD-stack} achieves an overall answer accuracy of $97.02\%$, $54.25\%$, and $49.14\%$ on the longest dialogs ($L = 30$) that are created from scenes with $10$, $15$, and $20$ objects, respectively.

\begin{table}[!h]
    \begin{minipage}{1\linewidth}
      \centering
        \scalebox{0.7}[0.7]{
            \begin{tabular}{cl||cc||cc}
            \hline
            & \multirow{2}*{\textbf{Model}} &\multicolumn{2}{c||}{\textbf{Without FT}} & \multicolumn{2}{c}{\textbf{FT on split BB}} \\
            \cmidrule(r){3-4} \cmidrule(r){5-6}
            & &\textbf{Acc.}& $\mathbf{{NFFR} \uparrow}$&\textbf{Acc.}& $\mathbf{{NFFR} \uparrow}$\\
            \hline
            \multirow{6}*{\rotatebox{90}{\textbf{Hist. + GT}}}  
            &\multicolumn{1}{|l||}{MAC-CQ}       & $36.12^\ddag$  & $0.14$\,\,  & $40.33^\dag$ & $0.16$\,\,\\
            &\multicolumn{1}{|l||}{$\quad$+ CAA} & $35.09^\dag$     & $0.14$\,\,  & $40.36^*$    & $0.16$\,\,\\
            &\multicolumn{1}{|l||}{$\quad$+ MTM} & $34.73$\,\,       & $0.15$\,\,  & $35.09$\,\,      & $0.13$ \,\,\\
            \cline{2-6}
            &\multicolumn{1}{|l||}{HCN}          & $47.67^\ddag$         & $0.14$\,\,  & $70.43^\ddag$     & $0.27\,\,$ \\
            &\multicolumn{1}{|l||}{\cellcolor{gray!25}NSVD-concat}& \cellcolor{gray!25}$64.07^\ddag$ & \cellcolor{gray!25}$0.24$\,\, &  \cellcolor{gray!25}$99.44^\ddag$ & \cellcolor{gray!25}$0.96$\,\,\\
            &\multicolumn{1}{|l||}{\cellcolor{gray!25}NSVD-stack} &
            \cellcolor{gray!25}{$\mathbf{71.55}^\ddag$} & \cellcolor{gray!25}{$\mathbf{0.28}$}\,\, &  \cellcolor{gray!25}{$\mathbf{99.51}^\dag$} & \cellcolor{gray!25}{$\mathbf{0.97}$}\,\,\\ 
            \hline
            \multirow{6}*{\rotatebox{90}{\textbf{Hist. + Pred.}}}  
            
            &\multicolumn{1}{|l||}{MAC-CQ}        & $35.09^\ddag$ & $0.13$\,\,  &  $39.53^\ddag$ & $0.15$\,\, \\
            &\multicolumn{1}{|l||}{$\quad$+ CAA}  & $36.40^\ddag$ & $0.14$\,\,  &  $37.72^\ddag$ & $0.15$\,\,\\
            &\multicolumn{1}{|l||}{$\quad$+ MTM}  & $6.19$\,\,       &  $0.09$\,\, &  $7.03$\,\,       & $0.09$\,\,\\
            \cline{2-6}

            &\multicolumn{1}{|l||}{HCN}           & $46.91^\ddag$         & $0.13$\,\,  & $68.82^\ddag$    & $0.25$\,\, \\

            &\multicolumn{1}{|l||}{\cellcolor{gray!25}NSVD-concat} & \cellcolor{gray!25}$64.07^\ddag$ & \cellcolor{gray!25}$$0.24$$\,\, & \cellcolor{gray!25}$99.44^\ddag$ & \cellcolor{gray!25}$0.96$\,\,\\
            
            &\multicolumn{1}{|l||}{\cellcolor{gray!25}NSVD-stack} & \cellcolor{gray!25}{$\mathbf{71.55}^\ddag$} & \cellcolor{gray!25}{$\mathbf{0.28}$}\,\, &  \cellcolor{gray!25}{$\mathbf{99.51}^\dag$} & \cellcolor{gray!25}{$\mathbf{0.97}$}\,\,\\ 
            \hline
            \end{tabular}
        }
        \caption{Results when training on split AA and testing on split BB. 
        Best results in bold.
        $\ddag, \dag$, and $*$ represent $p < 0.00001$, $p < 0.05$ and $p \geq 0.05$ compared to the second best score in each column, respectively.}
    \label{tab:AB_splits}
    \end{minipage} 
\end{table}

\paragraph{Generalisation to Unseen Questions Types.}
Similar to prior works \cite{Johnson2017,nsvqa,Mao2019}, we addressed the generalisability to new scenes and object combinations in our previous experiments.
However, generalisability to unseen questions remains unexplored. 
To address this, we created two splits (AA and BB) on CLEVR-Dialog as follows:
we first split the CLEVR validation images into two disjoint halves A and B.
We then split the question types into split A and split B.
We randomly assigned half of the question types in each category to split A and the other half to split B to prevent biasing either one to a particular question category.
For each image in both splits, we generated five dialogs consisting of $10$ rounds as in \cite{Kottur2019}. 
Split AA contains a training and a validation set based on $6,000$ and $1500$ images, respectively. 
Split BB has a fine-tuning, a validation, and a test set generated on $2000$, $500$, and $5000$ images, respectively.
The desired behaviour for a model that generalises well is to perform well on split BB when only trained on split AA.

\textit{NSVD-concat} and \textit{NSVD-stack} achieve an accuracy of $64.07\%$ and $71.55\%$, respectively, when tested on split BB without fine-tuning, thereby significantly outperforming all baselines (\autoref{tab:AB_splits}).
However, low $\textrm{NFRR}$ values
indicate that first failures occur early on in the dialog.
After fine-tuning all models on a small amount of data from split BB,
our models achieve accuracies and $\textrm{NFRR}$ values comparable to previous experiments.
In stark contrast, purely-connectionist baselines' performance 
only improves by a small margin with the highest jump of $5.26\%$ being achieved by \textit{MAC-CQ-CAA}.
This shows an impressive data efficiency of the neuro-symbolic models that, in contrast to the data-hungry purely-connectionist baselines, are able to learn and adapt efficiently from a very small amount of data.
More details regarding the training data efficiency can be found in the Appendix \ref{sec:data_efficiency}.

    \section{Conclusion and Future Work}
We proposed \methodName \textemdash the first hybrid method to combine deep learning and symbolic program execution for multi-round visual reasoning \textemdash and a new, stricter and more realistic evaluation scheme
for visual dialog.
Our method outperforms state-of-the-art purely-connectionist baselines on CLEVR-Dialog and sets a new near-perfect test accuracy of $99.72\%$. 
Furthermore, \methodNameShort has a higher mean First Failure Round, is more robust to incomplete dialog histories, and generalises better to longer dialogs and to unseen question types and scenes.
These performance improvements are not to be seen in isolation of the stricter supervision our models have as they require a supervised fine-tuning of a Mask-RCNN in addition to a supervised training of the program parsers.   
Finally, additional evaluations show that our models generalise to other scene domains (Appendix \ref{sec_minecraft}) and we expect them to even generalise to naturalistic datasets as well if they provide the necessary supervision requirements for our models similar to the VQA scenario \cite{Wang2017,Gan2017}. To the best of our knowledge, such datasets for visual dialog do not yet exist.

    \section*{Acknowledgments}
We would like to thank the anonymous reviewers for their insightful comments, Jiayuan Mao for providing us with the Minecraft data, and Dominike Thomas for paper editing support. M. Bâce was funded by a Swiss National Science Foundation (SNSF) Early Postdoc Mobility Fellowship (No. 199991). A. Bulling was funded by the European Research Council (ERC; grant agreement 801708).

    \bibliography{main}
    \bibliographystyle{acl_natbib}
    
    \appendix
    
\section{Appendix}
\begin{sidewaystable*}
\centering
\scalebox{1}[1]{
\begin{tabular}{clccccccc}
\hline
& \multirow{2}*{\textbf{Func. Name}} & \multirow{2}*{\textbf{Func. Args.}} & \multirow{2}*{\textbf{Func. Out.}}  & \multirow{2}*{\textbf{\texttt{fetch}}} & \multicolumn{4}{c}{\textbf{\texttt{update}}}\\

\cmidrule(r){6-9}

&&&&&\textbf{\texttt{Handle}}&\textbf{\texttt{Conv.\,\,Subj.}}&\textbf{\texttt{Seen\,\,Objs.}}&\textbf{\texttt{Groups}}\\
\hline

\multirow{8}*{\rotatebox{90}{\textbf{Caption Programs}}}
&\multicolumn{1}{|l}{\texttt{count-att}}     & \texttt{attr} & \texttt{none} & \xmark &\cmark & \xmark & \cmark & \cmark\\
&\multicolumn{1}{|l}{\texttt{extreme-right}} & $\uplus$ \texttt{[attr\_1,..,attr\_4]}  & \texttt{none} & \xmark &\cmark &\cmark &\cmark&\xmark \\
&\multicolumn{1}{|l}{\texttt{extreme-left}}  &  $\uplus$ \texttt{[attr\_1,..,attr\_4]} & \texttt{none}& \xmark &\cmark &\cmark &\cmark&\xmark \\
&\multicolumn{1}{|l}{\texttt{extreme-behind}}&  $\uplus$ \texttt{[attr\_1,..,attr\_4]} & \texttt{none}& \xmark &\cmark &\cmark &\cmark&\xmark \\
&\multicolumn{1}{|l}{\texttt{extreme-front}} & $\uplus$ \texttt{[attr\_1,..,attr\_4]}  & \texttt{none}& \xmark &\cmark &\cmark &\cmark&\xmark \\
&\multicolumn{1}{|l}{\texttt{extreme-centre}}&  $\uplus$ \texttt{[attr\_1,..,attr\_4]} & \texttt{none} & \xmark &\cmark &\cmark &\cmark&\xmark\\
&\multicolumn{1}{|l}{\texttt{unique-obj}}    & $\uplus$ \texttt{[attr\_1,..,attr\_4]}  & \texttt{none}& \xmark &\cmark &\cmark &\cmark&\xmark\\
&\multicolumn{1}{|l}{\texttt{obj-relation}}  & \texttt{attr\_obj\_1,pos,attr\_obj\_2} & \texttt{none}& \xmark &\cmark &\cmark &\cmark&\xmark\\

\hline
\multirow{24}*{\rotatebox{90}{\textbf{Question Programs}}}  
&\multicolumn{1}{|l}{\texttt{count-all}}  & \texttt{-} & \texttt{num}& \xmark &\xmark &\xmark &\xmark&\cmark \\
&\multicolumn{1}{|l}{\texttt{count-other}}   & \texttt{-} & \texttt{num}& \xmark  &\xmark &\cmark &\cmark &\xmark \\
&\multicolumn{1}{|l}{\texttt{count-all-group}}  & \texttt{-} & \texttt{num} & \xmark  &\xmark &\xmark &\xmark&\xmark\\
&\multicolumn{1}{|l}{\texttt{count-attribute}}  & \texttt{attr} &\texttt{num}& \xmark  &\cmark &\cmark &\cmark&\cmark\\
&\multicolumn{1}{|l}{\texttt{count-attribute-group}}  & \texttt{attr} &\texttt{num}& \xmark &\cmark &\cmark &\cmark&\cmark\\
&\multicolumn{1}{|l}{\texttt{count-obj-rel-imm}}  & \texttt{pos} & \texttt{num}& \xmark &\xmark &\cmark &\cmark&\cmark\\
&\multicolumn{1}{|l}{\texttt{count-obj-rel-imm-2}}  &\texttt{pos} & \texttt{num}& \xmark  &\xmark &\cmark &\cmark&\cmark\\
&\multicolumn{1}{|l}{\texttt{count-obj-rel-early}}  & \texttt{pos,attr} & \texttt{num}& \cmark  &\cmark &\cmark &\cmark&\cmark\\
&\multicolumn{1}{|l}{\texttt{count-obj-exclude-imm}}  & \texttt{attr\_type} & \texttt{num}& \xmark  &\xmark &\cmark &\cmark&\cmark\\
&\multicolumn{1}{|l}{\texttt{count-obj-exclude-early}}  & \texttt{attr\_type,attr} & \texttt{num} & \cmark  &\xmark &\cmark &\cmark&\cmark\\

&\multicolumn{1}{|l}{\texttt{exist-other}}  & \texttt{-} & \texttt{yes/no} & \xmark  &\xmark &\xmark &\cmark&\cmark \\
&\multicolumn{1}{|l}{\texttt{exist-attribute}}  & \texttt{attr} & \texttt{yes/no} & \xmark  &\cmark &\xmark &\cmark&\cmark\\
&\multicolumn{1}{|l}{\texttt{exist-attribute-group}}  & \texttt{attr} & \texttt{yes/no} & \xmark &\cmark &\cmark &\cmark&\cmark \\
&\multicolumn{1}{|l}{\texttt{exist-obj-rel-imm}}  & \texttt{pos} & \texttt{yes/no} & \xmark &\xmark &\cmark &\cmark&\cmark\\
&\multicolumn{1}{|l}{\texttt{exist-obj-rel-imm2}} & \texttt{pos} & \texttt{yes/no} & \xmark &\xmark &\cmark &\cmark&\cmark\\
&\multicolumn{1}{|l}{\texttt{exist-obj-rel-early}}  & \texttt{pos,attr} & \texttt{yes/no} & \cmark &\cmark &\cmark &\cmark&\cmark\\
&\multicolumn{1}{|l}{\texttt{exist-obj-exclude-imm}}  & \texttt{attr\_type} & \texttt{yes/no}& \xmark &\xmark &\cmark &\cmark&\cmark\\
&\multicolumn{1}{|l}{\texttt{exist-obj-exclude-early}}  & \texttt{attr\_type,attr} & \texttt{yes/no}& \cmark &\xmark &\xmark &\xmark&\cmark\\

&\multicolumn{1}{|l}{\texttt{seek-attr-imm}}  & \texttt{attr\_type} & \texttt{attr}& \xmark &\cmark &\xmark &\xmark&\xmark\\
&\multicolumn{1}{|l}{\texttt{seek-attr-imm2}}  & \texttt{attr\_type} & \texttt{attr}& \xmark&\cmark &\xmark &\xmark&\xmark\\
&\multicolumn{1}{|l}{\texttt{seek-attr-early}}  & \texttt{attr\_type,attr} & \texttt{attr}& \cmark &\cmark &\cmark &\cmark&\xmark\\
&\multicolumn{1}{|l}{\texttt{seek-attr-sim-early}}  & \texttt{attr\_type,attr} & \texttt{attr}& \cmark &\cmark &\cmark &\cmark&\xmark\\
&\multicolumn{1}{|l}{\texttt{seek-attr-rel-imm}}  & \texttt{attr\_type} & \texttt{attr}& \xmark &\cmark &\cmark &\cmark&\xmark\\
&\multicolumn{1}{|l}{\texttt{seek-attr-rel-early}}  & \texttt{attr\_type,pos,attr} & \texttt{attr}& \cmark &\cmark &\cmark &\cmark&\xmark\\

\hline
\end{tabular}
}
\caption{Our Domain Specific Language (DSL) for CLEVR-Dialog. We present each function with its expected argument types, output, and knowledge base operations. 
Given a set of $n$ arguments, the $\uplus$ operator selects a subset of $m\leq n$ possible ones, e.g. $\uplus \texttt{[attr\_1,..,attr\_4]} = \texttt{[attr\_2,attr\_3]}.$
}
\label{tab:dsl}
\end{sidewaystable*}
\begin{table*}
$\texttt{attr\_1} \in \texttt{COLOURS=[blue,brown,cyan,grey,green,purple,red,yellow],}$\\
$\texttt{attr\_2} \in \texttt{MATERIALS=[rubber,metal],}$ \\
$\texttt{attr\_3} \in \texttt{SHAPES=[cube,cylinder,sphere],}$ \\
$\texttt{attr\_4} \in \texttt{SIZES=[large,small],}$ \\
$\texttt{attr}, \texttt{attr\_obj\_1}, \texttt{attr\_obj\_2} \in \bigcup \texttt{\{COLOURS,MATERIALS,SHAPES,SIZES\},}$\\
$\texttt{attr\_type} \in \texttt{[colour,material,shape,size],}$\\
$\texttt{pos} \in \texttt{[right,left,front,behind],}$\\
$\texttt{num} \in \texttt{[0,1,2,3,4,5,6,7,8,9,10].}$\\
\protect\caption{Argument types of our DSL.}
\label{tab:arg_types}
\end{table*}

\subsection{The CLEVR-Dialog DSL}
\label{sec:appendix_dsl}
Our Domain Specific Language (DSL) for CLEVR-Dialog is depicted in \autoref{tab:dsl}. We present each function with its expected argument types, output, and knowledge base operations.
The argument types are further defined in \autoref{tab:arg_types}.
We use the variables \texttt{attr}, \texttt{attr\_obj\_1}, \texttt{attr\_obj\_2} and \texttt{attr\_i} for $\texttt{i = 1,..,4}$ to denote the set of possible CLEVR attributes, i.e. colour, material, shape, and size. Furthermore, the variable \texttt{pos} denote one possible position, i.e. \texttt{right}, \texttt{left}, \texttt{front}, or \texttt{behind}.
Finally, the variable \texttt{num} denotes the set of possible numerical values , i.e. between 0 and $N$, where $N$ is the maximum number of objects in the scene. If not explicitly stated otherwise, we assume $N=10$. 
\subsection{Normalised First Failure Round}
\label{sec:appendix_ffr}
The Normalised First Failure Round ${\textrm{NFFR}}$ is calculated as $${\textrm{NFFR}} = \displaystyle{\frac{1}{N}}\sum_{i=1}^N \displaystyle{\frac{1}{L+1}} \sum_{j=1}^L \Delta_j^{(i)} \delta_{\textrm{pred}_j^{(i)}, \textrm{gt}_j^{(i)}} \alpha^{(i)}_j,$$
where $N$ is the total number of dialogs, $L$ is the length of each dialog, and $\textrm{pred}_j^{(i)}$ and $\textrm{gt}_j^{(i)}$ are the predicted and ground truth answers at round $j$ of dialog $i$, respectively. 
Furthermore, for each round $j$ of every dialog $i$, we define  $\Delta_j^{(i)}$,  $\delta_{\textrm{pred}_j^{(i)}, \textrm{gt}_j^{(i)}}$, and $\alpha^{(i)}_j$ as:
\[ \Delta_j^{(i)} = 
	\left\{
	\begin{array}{ll}
	j \quad \textrm{if} \quad j \leq L\\
	L+1 \quad \textrm{if} \quad j=L \wedge   \delta_{\textrm{pred}_j^{(i)}, \textrm{gt}_j^{(i)}} = 1
	\end{array}
	\right.
,\]

\[ \delta_{\textrm{pred}_j^{(i)}, \textrm{gt}_j^{(i)}} = 
	\left\{
	\begin{array}{ll}
	1 \quad \textrm{if} \quad \textrm{pred}_j^{(i)} \neq \textrm{gt}_j^{(i)}\\
	0 \quad \textrm{otherwise}
	\end{array}
	\right.
,\]
\[ \alpha_j^{(i)} = 
	\left\{
	\begin{array}{ll}
	0 \quad \textrm{if} \,\,\, \exists k<j \,\,\, \textrm{s.t.} \,\,\, \delta_{\textrm{pred}_k^{(i)}, \textrm{gt}_k^{(i)}} = 1 \\
	1 \quad \textrm{otherwise}
	\end{array}
	\right.
.\] 

By definition, we set the NFFR of a model to be $L+1$ if it correctly answers all $L$ dialog rounds.

\subsection{Training Details}
\label{sec:tr_details}
\paragraph{Our Models.}
In order to encode our raw data, we generated two different vocabularies from the training data: one that handles the captions, questions, and answers in form of natural language and another that deals with ground truth caption and question programs. 
For encoding, we pad all captions, questions, and programs to a maximum length of $n_q = 21$, $n_c = 16$, and $n_y = 6$, respectively.
Then, the corresponding tokens are transformed into a $300$-dim. space using either the encoder text embedding or the decoder program embedding.
We trained the caption and question program generators separately.
We used a $2$-layered bi-directional LSTM with a hidden size of $256$ in both caption and question encoders.
In the decoder, we used a $2$-layered LSTM with a hidden size of $512$.
We fixed the batch size to $64$ and used the Adam optimiser \cite{kingma2014adam} with a learning rate of $7\times10^{-4}$ to train for \textit{one} epoch. Every $2000$ iteration, we validated the models based on the program accuracy in order not to select one that follows a flawed logic, i.e. a wrong program, to predict an answer.

\paragraph{Baselines.}
We trained the purely-connectionist baselines using the official code
\footnote{\url{https://github.com/ahmedshah1494/clevr-dialog-mac-net/tree/dialog-macnet}} we obtained from the authors of \citet{shah-etal-2020-reasoning}.
We used the same hyperparameters and training scheme, i.e. we trained the models for a maximum of $25$ epochs and used early stopping when the validation accuracy did not improve for five consecutive epochs.
For the HCN models, we used a publicly available codebase\footnote{\url{https://github.com/jojonki/Hybrid-Code-Networks}} for training with the same hyper-parameters as in \cite{Williams2017}.

\begin{figure}[!t]
\begin{minipage}{1\linewidth}
        \centering
            \scalebox{1}[1.0]{
                \includegraphics[width=\textwidth]{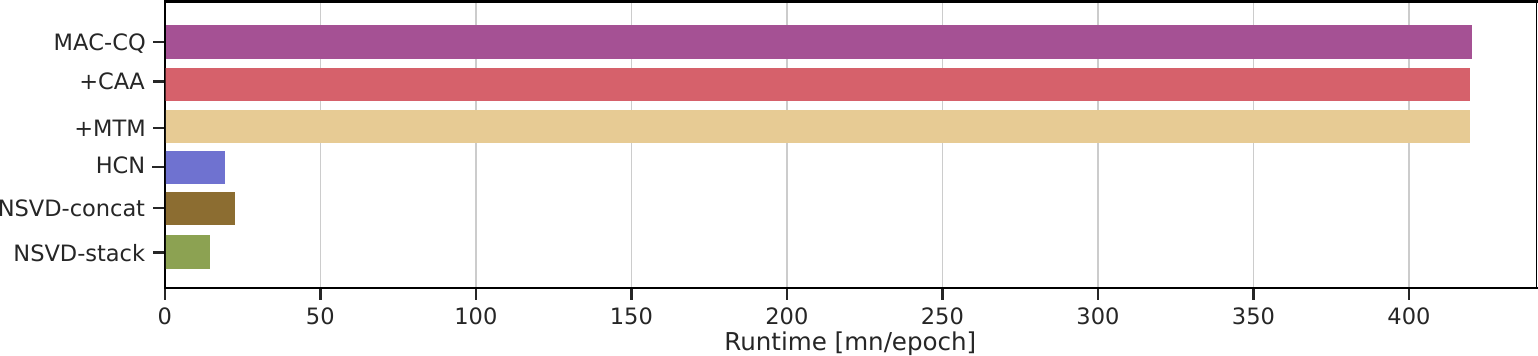}
            }
            \caption{
           Training runtime comparison of the models. The training was conducted on a single NVIDIA Tesla V100 GPU with batch size 64.}
            \label{fig:runtime}
    \end{minipage}%
\end{figure}

\paragraph{Runtime Analysis.}
We conducted all of our experiments on a single NVIDIA Tesla V100 GPU. 
The training runtimes for one epoch when using a batch size of $64$ are illustrated in \autoref{fig:runtime}.
As can be seen, all the purely-connectionist baselines need more than seven hours to complete one epoch.
Although NSVD-concat concatenates the previous dialog rounds to form the history similarly to the baselines,
it is more computationally efficient as it only needs circa 22 minutes to complete one epoch.
Finally, HCN needs around $19$ minutes to complete one epoch and NSVD-stack only around $14$ which solidifies our aforementioned efficiency claims.

\begin{figure*}[!t]
\begin{minipage}{1\linewidth}

        \centering
        \scalebox{1}[1]{
            \includegraphics[width=\textwidth]{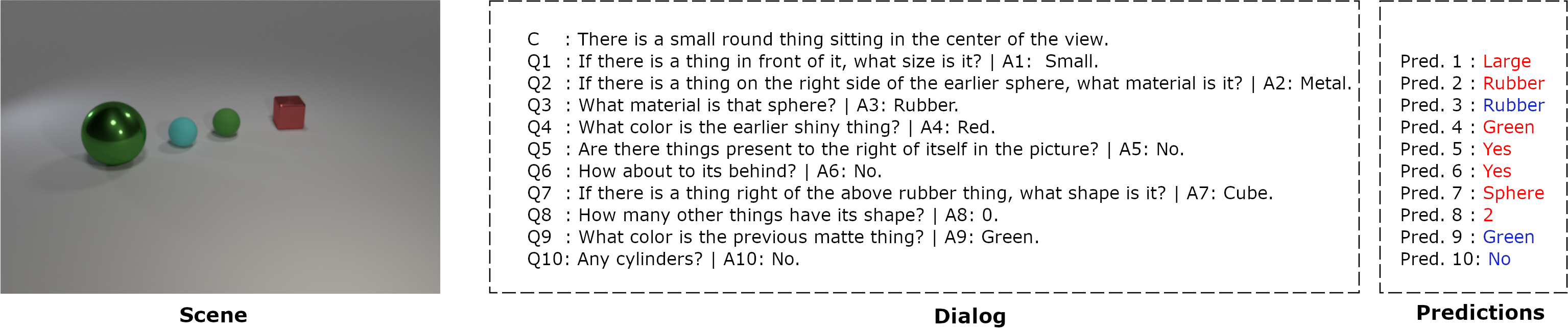}
        }
        \caption{Example of a situation where a caption cannot be uniquely interpreted. The program \texttt{extreme-centre(cylinder, small)} induced from the caption ``there is a small round thing sitting in the centre of the view'' does not lead to a unique initialisation of our executor's knowledge base as there are \textit{two} small spheres in the centre of the scene. By our logic,  we consider it to be the cyan one. Incorrectly initialising the knowledge base leads to confusion when answering the subsequent questions. The blue and red colours indicate a match or a mismatch between the predicted answer and the ground truth, respectively.}
        \label{fig:non_uniqueness}
\end{minipage}
\end{figure*}

\begin{table}
\centering
\scalebox{0.93}[0.93]{

\begin{tabular}{lccc}
\hline
\multirow{2}*{\textbf{Model}}& \multicolumn{2}{c}{\textbf{Prog. Acc.}} & \multirow{2}*{\shortstack{\textbf{Executor}\\ \textbf{Acc.}}}\\
\cmidrule(r){2-3}

& \textbf{Caption}&  \textbf{Question}\\
\hline
Caption-Net& $99.79$  & \multicolumn{1}{c|}{-} & \multirow{3}*{$99.99$}  \\
NSVD-concat & - &\multicolumn{1}{c|}{$99.87$} \\
NSVD-stack  & - &\multicolumn{1}{c|}{$99.99$} \\ 
\hline
\end{tabular}
}
\caption{Quantitative analysis of our models' logic. The high program accuracies demonstrate that our models follow the implemented logic to predict the correct answer, i.e. they do not execute false programs that by chance might lead to a correct prediction. In addition, when tested with the ground truth scene annotations and programs, our executor reaches an answer-accuracy of $99.99\%$ showcasing its flawless logic.}
\label{tab:logic_analysis}
\end{table}

\subsection{Quantitative Analysis of (our) Logic.}
\label{sec:logic_analysis}
To quantify the logical capabilities of our models, we measured the caption and question program accuracies on the test split. 
As we can see from \autoref{tab:logic_analysis}, our model achieves $99.79\%$ caption program accuracy and our best question generator reaches the $99.99\%$ accuracy mark.
That is, they do not follow a flawed logic, i.e. a wrong program, to predict the correct answer.
Furthermore, we evaluated the logic of our program executor by measuring its answer accuracy when provided with the ground truth programs and scene annotations of the test split. 
The last column of \autoref{tab:logic_analysis} shows that it reaches $99.99\%$ answer accuracy indicating that its logic is close to flawless.
The reason why it does not reach the $100\%$ mark is that some scene captions cannot be uniquely interpreted leading to potential confusions when answering the subsequent questions.
\autoref{fig:non_uniqueness} illustrates a concrete example of such a case. The caption ``there is a small round thing sitting in the centre of the view'' induces the program \texttt{extreme-centre(cylinder, small)}. However, this can be interpreted in two different ways since there are \textit{two} small spheres in the centre of the scene, i.e. the cyan and the green ones.
Therefore, the performance of our executor at answering the following dialog questions depends on which object is considered as the central one.
By our logic, we consider it to be the cyan one. The subsequent questions, ground truth answers and predictions are also shown in \autoref{fig:non_uniqueness}.

\begin{figure}[!t]
\begin{minipage}{1\linewidth}
        \centering
        \scalebox{1.0}[1.0]{
            \includegraphics[width=\textwidth]{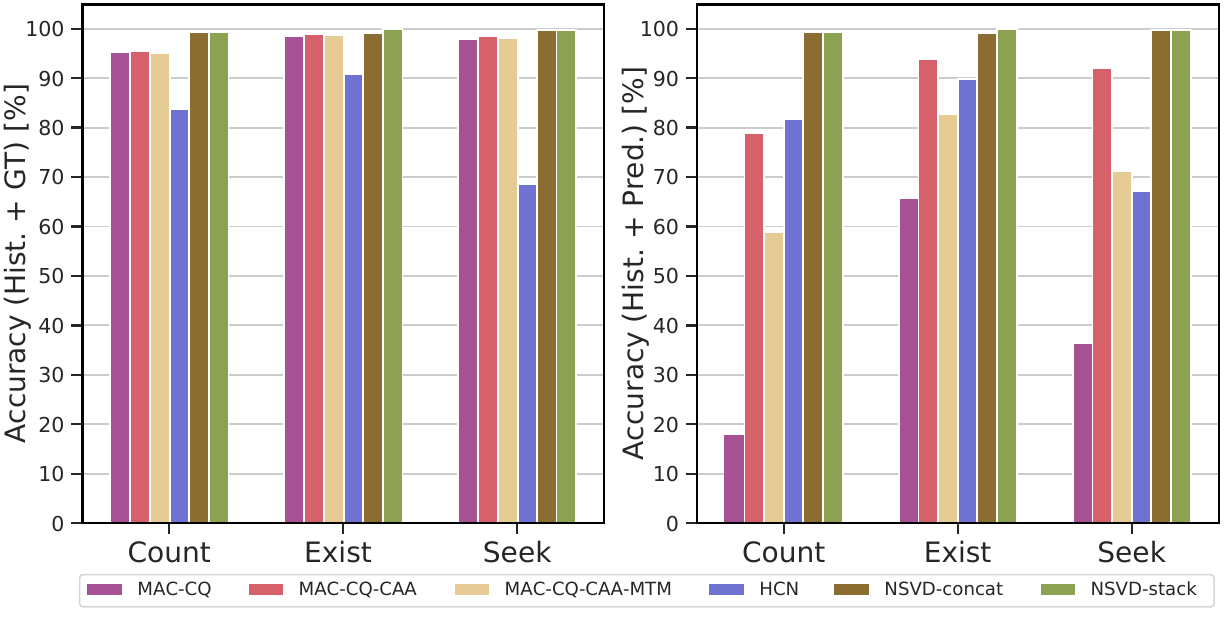}
        }
        \caption{Detailed performance comparison for the different question categories (\texttt{Count}, \texttt{Exist}, \texttt{Seek}). Independent of the evaluation approach, our models achieve new state-of-art results on all question categories. These improvements are statistically significant with $p<0.0001$ in all categories.}
        \label{fig:question_categories}
\end{minipage}%
\end{figure}

\begin{figure}[!t]
    \begin{minipage}{1\linewidth}
        \centering
            \scalebox{1}[1]{
                \includegraphics[width=\textwidth]{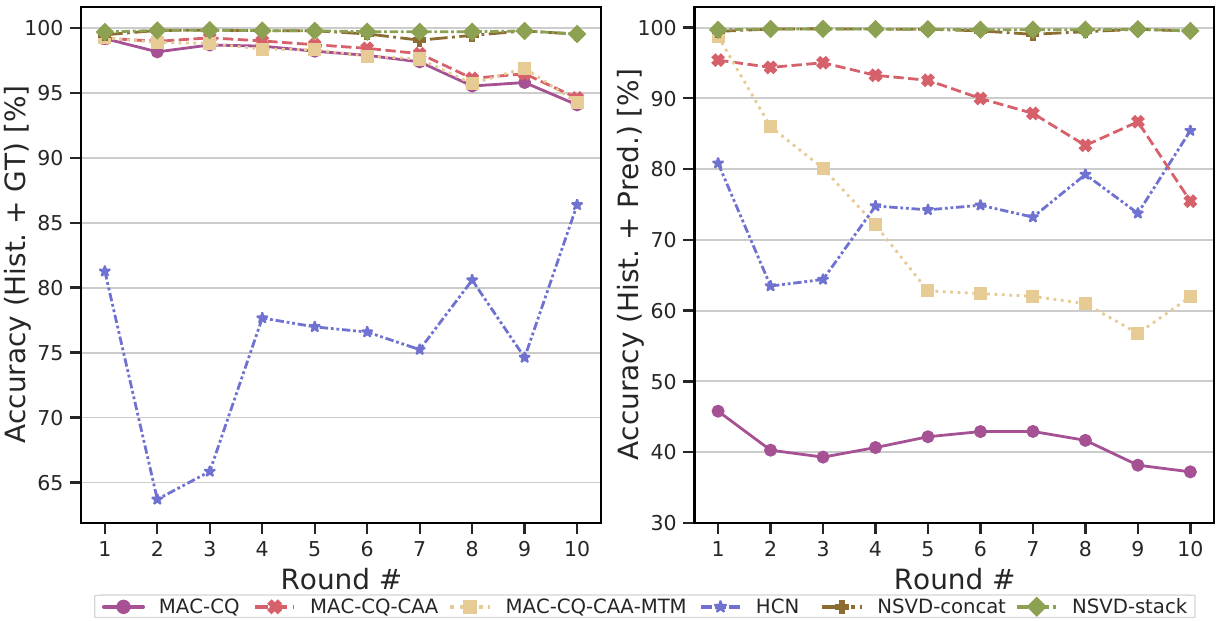}
            }
            \caption{
            Accuracy for different question rounds. 
            Our models  significantly outperform the baselines with $p<0.0001$ in all question rounds.}
            \label{fig:dialog_round}
    \end{minipage}%
\end{figure}

\begin{figure*}[!t]
        \centering
        \scalebox{1}[1]{
            \includegraphics[width=\textwidth]{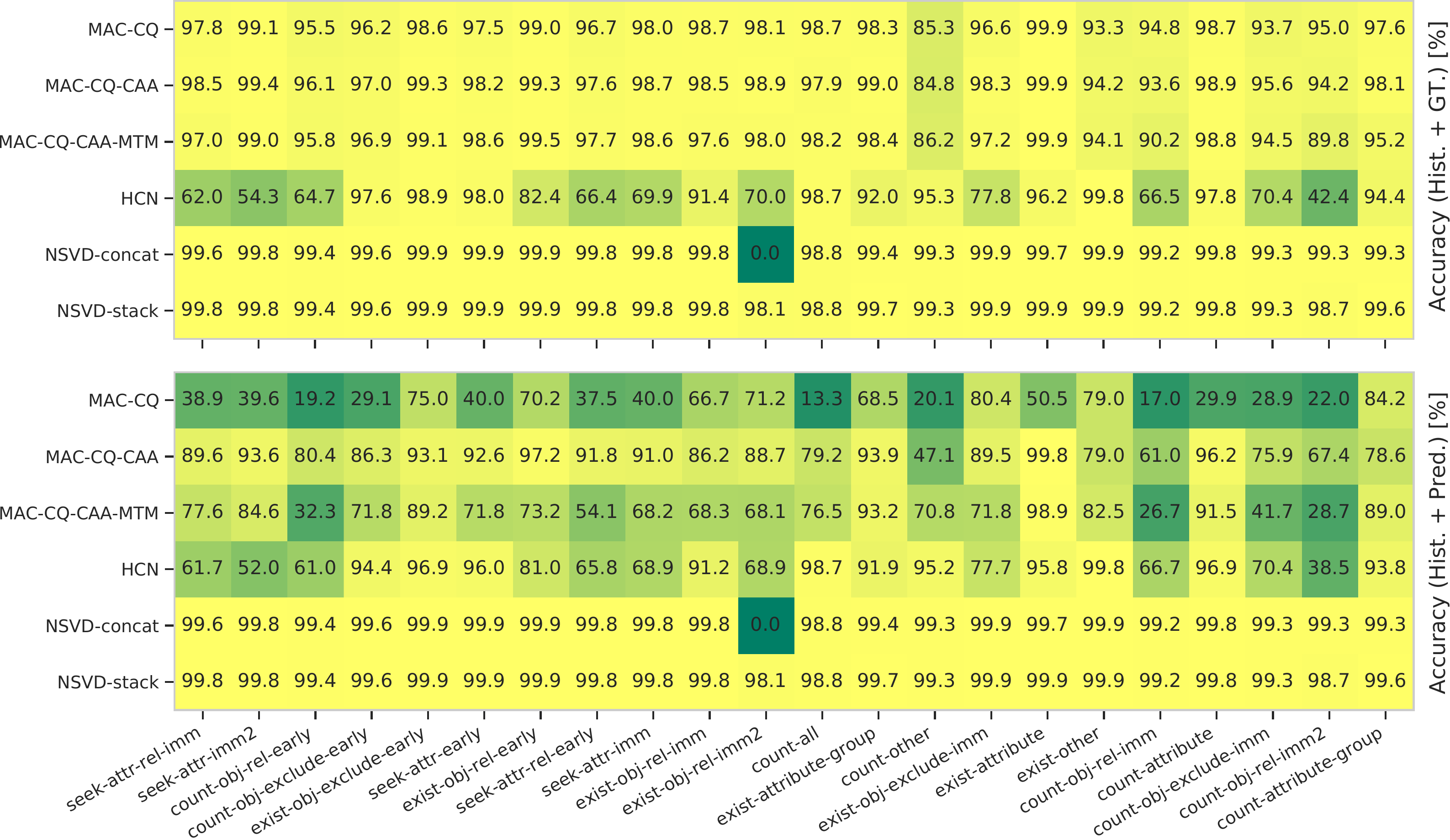}
        }
        \caption{Performance comparison on the individual question types. \textbf{Top:} Models were evaluated following the ``Hist. + GT'' evaluation scheme. \textbf{Bottom:} Models were evaluated following the ``Hist. + Pred.'' evaluation scheme }
        \label{fig:qtype_matrix}
\end{figure*}
\subsection{Fine-grained Evaluations}
\label{sec:finegrained_eval}
When looking at the performance for different question rounds (\autoref{fig:dialog_round}), we can see that while the performance of the baselines starts to deteriorate quickly with longer dialogs, our models achieve a consistently-high performance across all rounds.
The inability to maintain performance becomes even more apparent (especially for the purely-connectionist baselines) when using the stricter ``Hist. + Pred.'' evaluation scheme. 
The best baseline, \textit{MAC-CQ-CAA}, suffers from a drop of $8.48\%$ in accuracy and $0.19$ in ${\textrm{NFFR}}$.
The same observation can be made when analysing performance for the different CLEVR-Dialog question categories (\texttt{Count}, \texttt{Exist}, and \texttt{Seek}).
Our models outperform all baselines for all categories and both evaluation schemes (\autoref{fig:question_categories}). 
These improvements are statistically significant with $p<0.00001$.

\autoref{fig:qtype_matrix} illustrates the performance of our models on individual question types compared to the baselines. As can be seen, our models outperform the baselines on almost all question types with \textit{NSVD-stack} topping them all with an accuracy of over  $98\%$ for all types. As seen from previous experiments, the gap between our models and the baseline becomes ever more conspicuous when the ``Hist. + Pred.'' evaluation scheme is deployed (bottom table of \autoref{fig:qtype_matrix}).

\subsection{Data Efficiency}
\label{sec:data_efficiency}

\begin{figure}[!t]
    \begin{minipage}{1\linewidth}

        \centering
        \scalebox{0.99}[0.99]{
            \includegraphics[width=\textwidth]{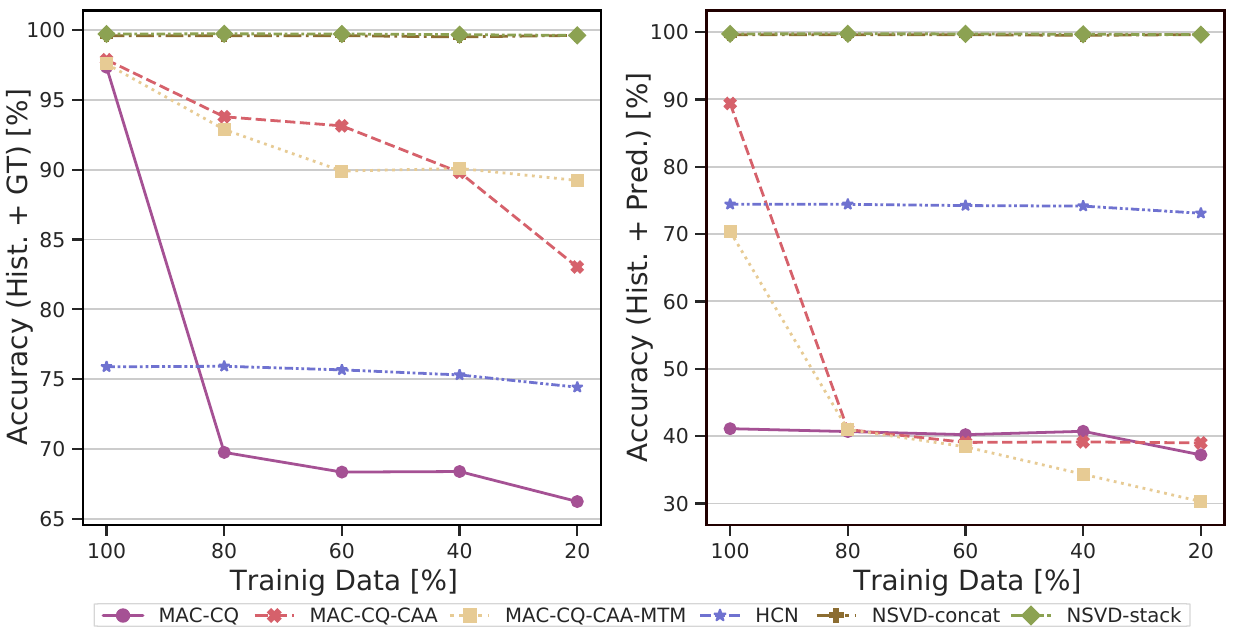}
        }
        \caption{
        Accuracy when trained on limited amounts of data ($20\%, 40\%, 60\%, 80\%,$ and $100\%$ of the overall training data) and evaluated on the test split following the ``Hist. + GT'' and ``Hist. + Pred.'' approaches.}
        \label{fig:data_efficiency}
    \end{minipage}
\end{figure}

To study the data efficiency of our models further, we trained them and the baselines on 
$20\%, 40\%, 60\%, 80\%,$ and $100\%$ of the available training data.
Not to bias the data towards any specific question type, we used the same distribution of question types to construct the reduced training sets.
After training, we evaluated the models on the test split using both evaluation schemes.
Our models not only outperform all baselines for the same reduced dataset but also in the extreme case of training the baselines with $100\%$ of the data and our models with only $20\%$ (\autoref{fig:data_efficiency}).
While the performance of the neuro-symbolic  models is only slightly affected by the size of the training data, the connectionist baselines' performance deteriorates with less data.
This deterioration becomes even more significant when the stricter ``Hist. + Pred.'' evaluation scheme is used.

\begin{figure*}[ht]
    \centering
    \scalebox{0.9}[0.9]{
    \includegraphics[width=0.25\linewidth]{
        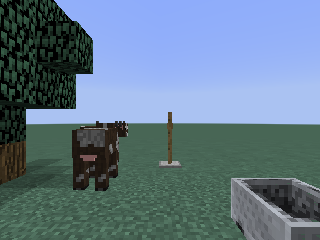}
    }\hfil
    \scalebox{0.9}[0.9]{
    \includegraphics[width=0.25\linewidth]{
        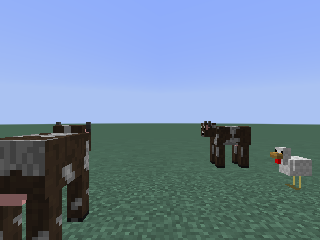}
    }\hfil
    \scalebox{0.9}[0.9]{
    \includegraphics[width=0.25\linewidth]{
        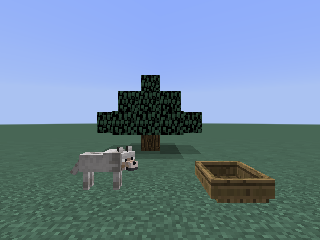}
    }\hfil
    \scalebox{0.9}[0.9]{
    \includegraphics[width=0.25\linewidth]{
        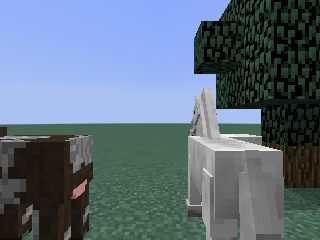}
    }\hfil
    \caption{Sample images from the Minecraft dataset.}
\label{fig:minecraft}
\end{figure*}

\subsection{Generalisation to Other Scene Domains}
\label{sec_minecraft}

\begin{table}[ht]
\centering
\scalebox{0.75}[0.75]{

\begin{tabular}{l||cc||cc}
\hline
\multirow{2}*{\textbf{Model}}& \multicolumn{2}{c||}{\textbf{Hist. + GT}} & \multicolumn{2}{c}{\textbf{Hist. + Pred.}}\\
\cmidrule(r){2-3} \cmidrule(r){4-5}
& \textbf{Acc.}&  $\mathbf{{NFFR} \uparrow}$& \textbf{Acc.}&  $\mathbf{{NFFR} \uparrow}$\\
\hline
MAC-CQ         & $64.30^*$       & $0.27$\,\,  & $59.96^\ddag$      & $0.24$\,\, \\
$\quad$+ CAA   & $64.28^\ddag$       & $0.27$\,\,  & $57.69^\ddag$  & $0.23$\,\, \\
$\quad$+ MTM   & $61.55^\ddag$       & $0.25$\,\,  & $52.04^\ddag$  & $0.20$\,\, \\
\hline
HCN            & $47.31\,\,$         & $0.14$\,\,  & $46.50\,\,$  & $0.14$\,\, \\
\rowcolor{Gray}
NSVD-concat & ${91.57}^\ddag$  & ${0.76}$\,\, & ${91.57}^\ddag$ & ${0.76}$\,\,  \\
\rowcolor{Gray}
NSVD-stack & $\mathbf{92.46^\ddag}$           &$\mathbf{0.83}$\,\,           & $\mathbf{92.46^\ddag}$          & $\mathbf{0.83}$\,\,           \\

\hline
\end{tabular}
}
\caption{Performance comparison on Minecraft-Dialog \textit{test}.
Results are shown for both ``Hist. + GT'' and ``Hist. + Pred.'' 
Our proposed models are highlighted in grey; best performance is in bold. $\ddag$ and $*$ represents $p < 0.00001$ and $p\geq 0.05$ compared to the second best score in the respective column, respectively.
}
\label{tab:overall_perfomance_minecraft}
\end{table}

In this experiment, we show that our method could be extended to a new reasoning testbed. Contrarily to CLEVR, the new scenes are grounded in Minecraft and are, as can be seen in \autoref{fig:minecraft}, drastically different in terms of context and scene constellations. Specifically, they comes with more entities (12 vs 3 in CLEVR) that have drastically different visual appearances. These entities can be grouped in a hierarchical manner (e.g. ``a cow'' and ``a wolf'' are both ``animals'' whereas ``an animal'' and ``a human'' are both ``creatures'').

Following \cite{nsvqa}, were rendered $10,000$ using the generation tool provided by Wu et al. \shortcite{nsd}. The scenes consist of three to six objects in a $2$D plane that are sampled from $12$ entities with $4$ different facing directions. Finally, we filtered out scenes that contain fully-occluded objects. We used $5,273$ images for training, $1,500$ for validation, and $1,000$ for testing. Furthermore, we adapted the dialog generation tool provided by Kottur et al. \shortcite{Kottur2019} to be able to account for the different scene properties.

Similar to previous experiments, we generated five dialogs for every image consisting of $10$ rounds each. We call this dataset Minecraft-Dialog.
The results are summarised in \autoref{tab:overall_perfomance_minecraft}. We compared our models to the same baselines as in previous experiments in terms of accuracy as well as $\textrm{NFRR}$. Once again, our neuro-symbolic models managed to significantly outperform all the baselines by achieving $92.64\%$ and $91.57\%$ accuracies for \textit{NSVD-stack} and \textit{NSVD-concat}, respectively, while maintaining high $\textrm{NFRR}$ values.

Contrarily, the best connectionist model achieved test accuracies of $64.30\%$ and $59.96\%$ using the ``Hist. + GT'' and ``Hist. + Pred.'' evaluation schemes, respectively.   
Finally, HCN achieved its best performance of $47.31$ accuracy and $0.14$ $\textrm{NFRR}$ when the ``Hist. + GT'' evaluation scheme was used.
Compared to CLEVR-Dialog (\autoref{tab:overall_perfomance}), the performance of our models witnessed a drop in this experiment which is attributed to the difficulty of the Minecraft scenes that come with heavy-occluded and diverse objects. However, the promising results of our models showcase that they indeed can generalise to new scene domains other then CLEVR.

\subsection{Input/Output Samples}
\label{sec:qualitative_ana} 
\begin{sidewaysfigure*}
\centering
\scalebox{1}[1]{
\includegraphics[width=\textwidth]{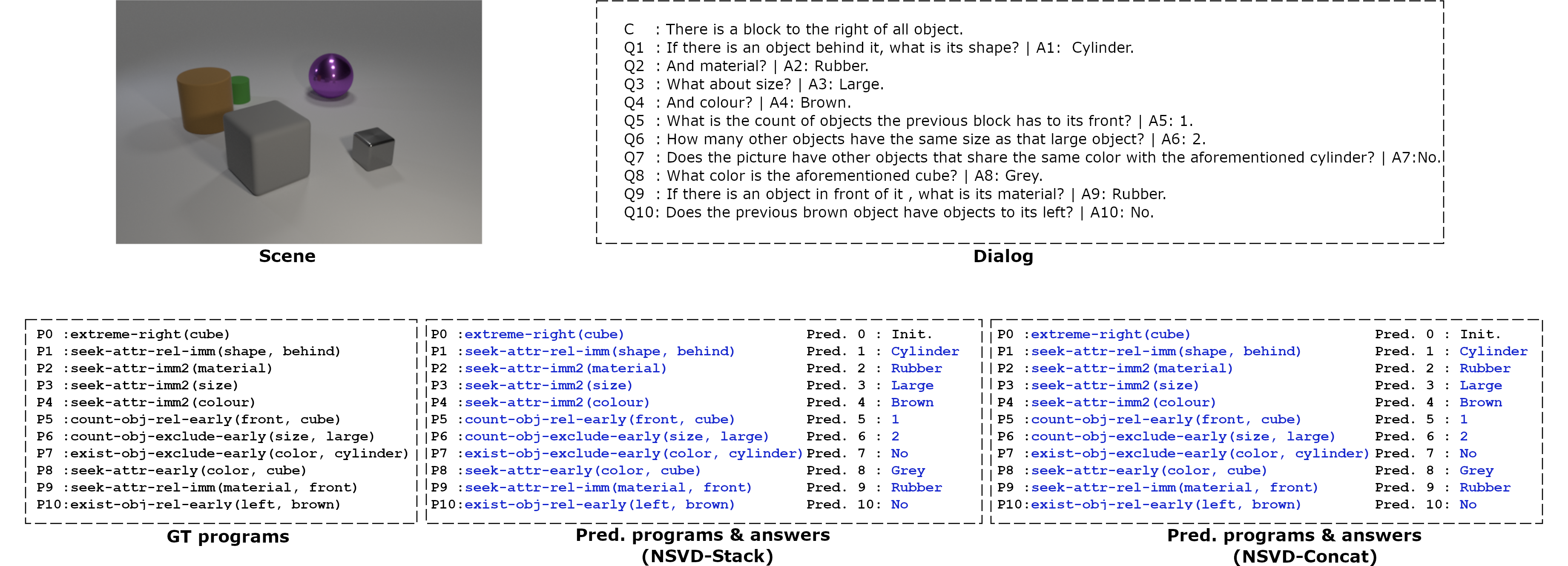}
}
\caption{Inference example of our models on a test instance. Both of our models executed correct programs to predict the answers. The blue colour indicates a match between the predicted program/answer and the ground truth.}
\label{fig:ex_1}
\end{sidewaysfigure*}

\begin{sidewaysfigure*}
\centering
\scalebox{1}[1]{
\includegraphics[width=\textwidth]{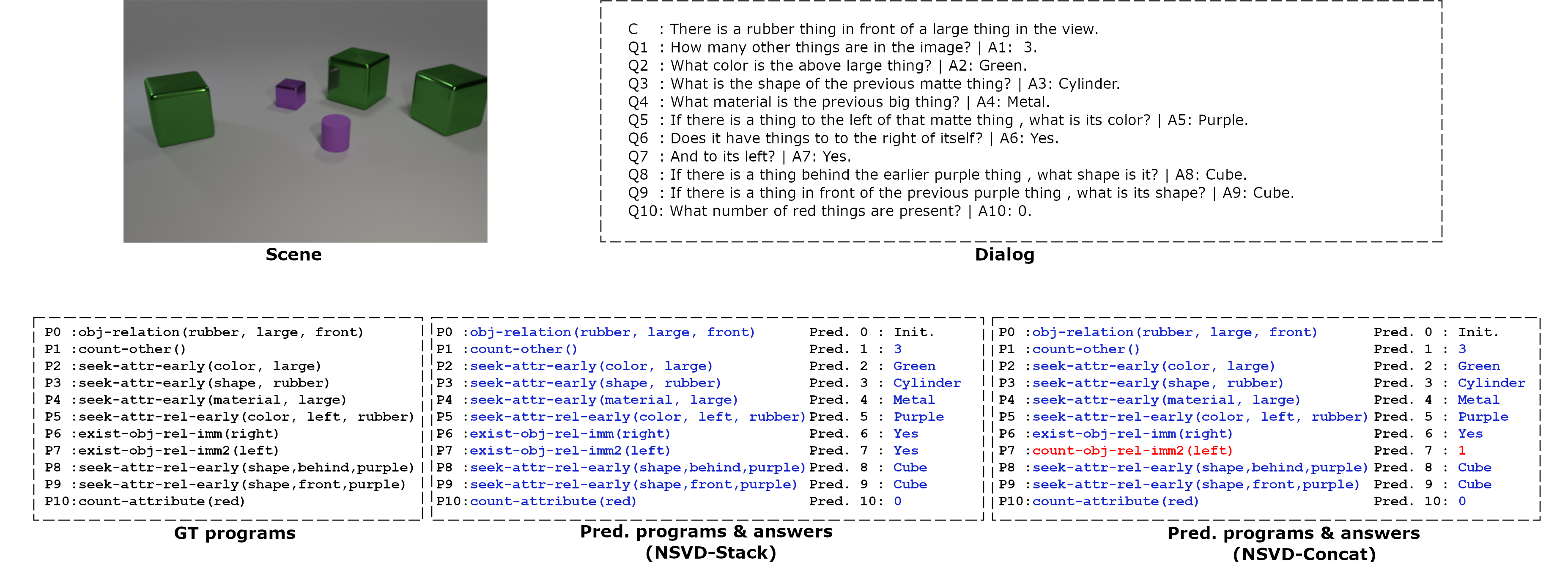}
}
\caption{Inference example of our models on a test instance. While \textit{NSVD-stack} answered all questions correctly, \textit{NSVD-concat} failed at round $7$. The blue and red colours indicate a match or a mismatch between the predicted program/answer and the ground truth, respectively.}
\label{fig:ex_2}
\end{sidewaysfigure*}

\begin{sidewaysfigure*}
\centering
\scalebox{1}[1]{
\includegraphics[width=\textwidth]{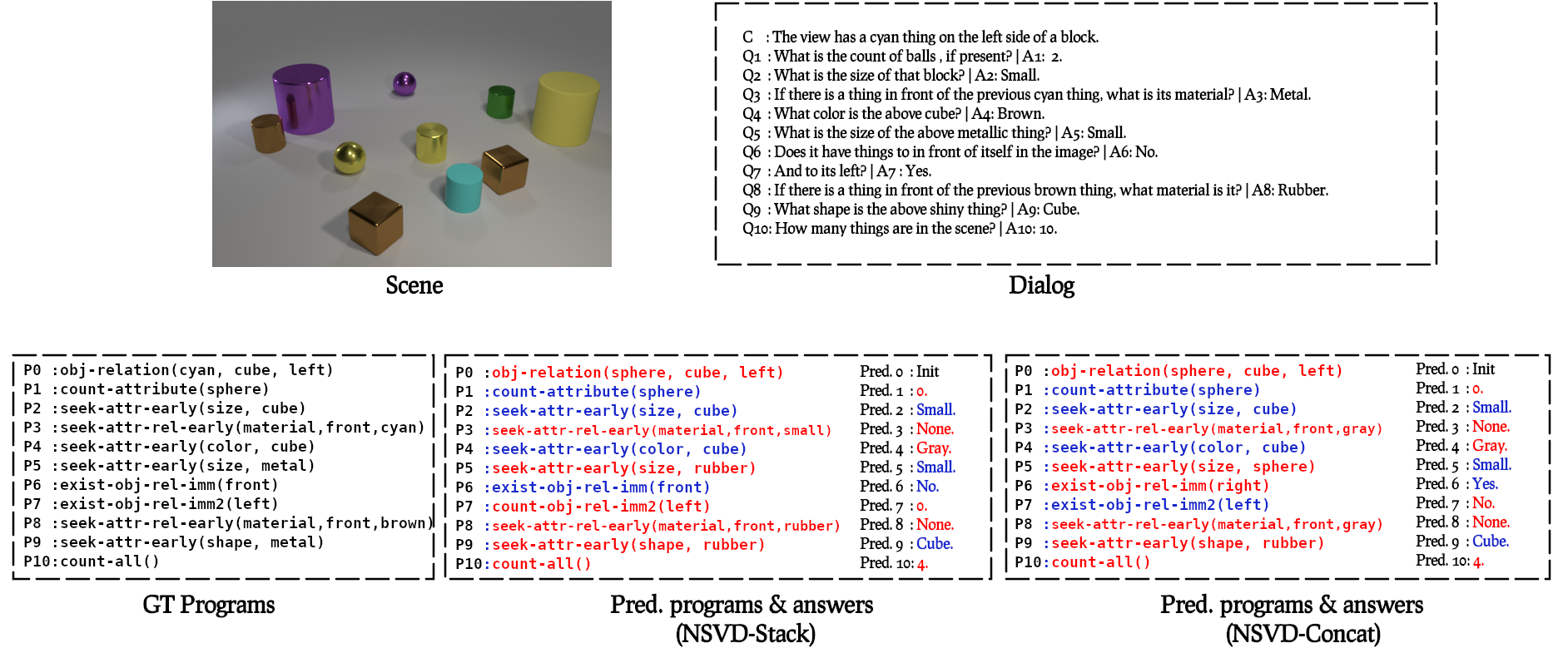}
}
\caption{Inference example of our models on a test instance with $10$ objects before fine-tuning. The blue and red colours indicate a match or a mismatch between the predicted program/answer and the ground truth, respectively.}
\label{fig:ex_3}
\end{sidewaysfigure*}

\begin{sidewaysfigure*}
\centering
\scalebox{1}[1]{
\includegraphics[width=\textwidth]{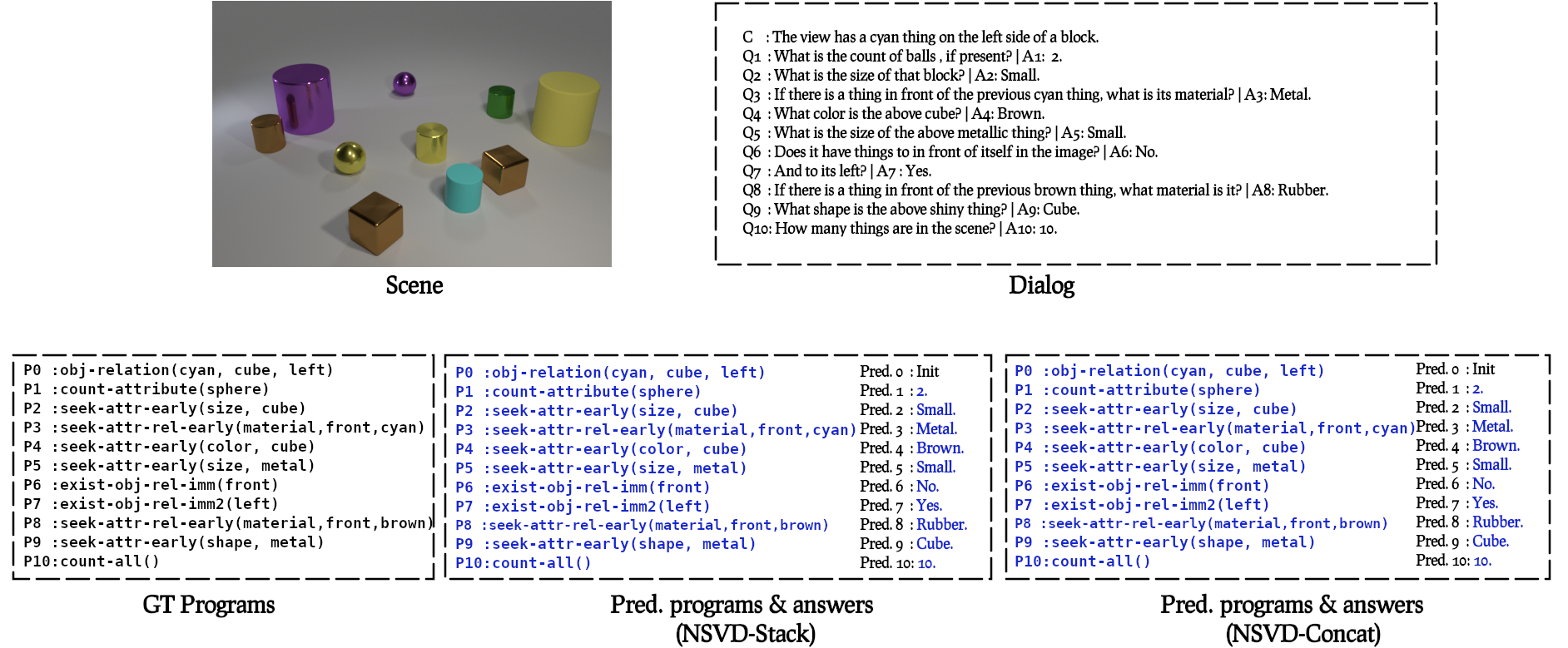}
}
\caption{Inference example of our models on a test instance with $10$ objects after fine-tuning. The blue and red colours indicate a match or a mismatch between the predicted program/answer and the ground truth, respectively.}
\label{fig:ex_4}
\end{sidewaysfigure*}
\begin{sidewaysfigure*}
\centering
\scalebox{1}[1]{
\includegraphics[width=\textwidth]{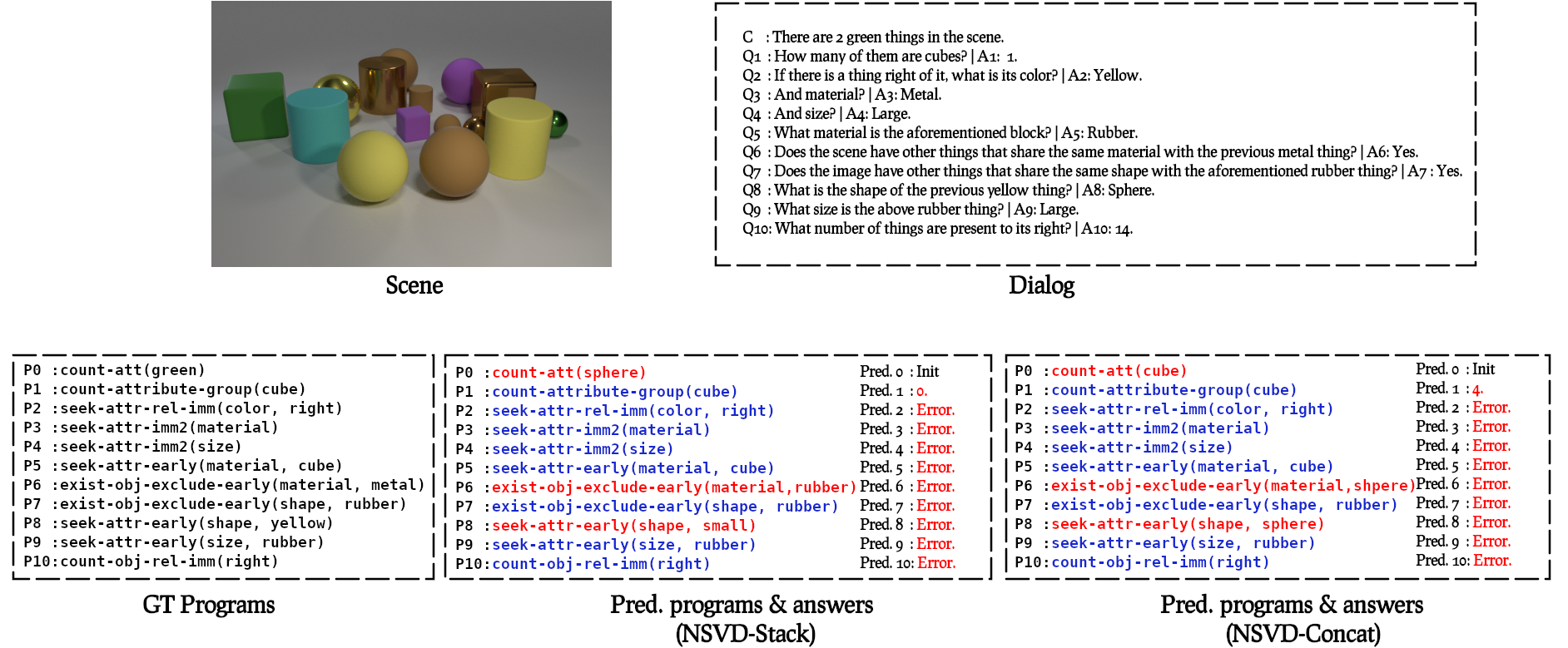}
}
\caption{Inference example of our models on a test instance with $15$ objects before fine-tuning. The blue and red colours indicate a match or a mismatch between the predicted program/answer and the ground truth, respectively. \texttt{Error} means that the predicted program encountered an exception during execution.}
\label{fig:ex_5}
\end{sidewaysfigure*}

\begin{sidewaysfigure*}
\centering
\scalebox{1}[1]{
\includegraphics[width=\textwidth]{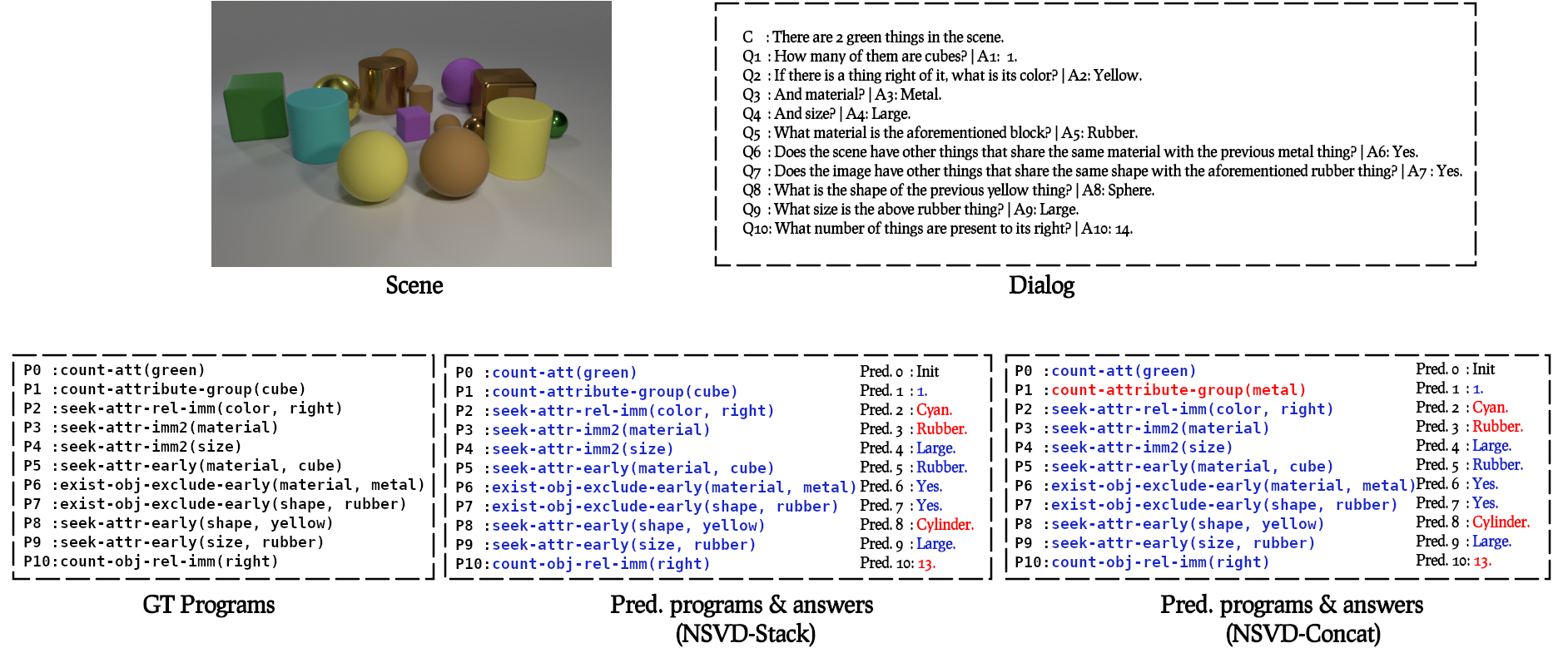}
}
\caption{Inference example of our models on a test instance with $15$ objects after fine-tuning. The blue and red colours indicate a match or a mismatch between the predicted program/answer and the ground truth, respectively.}
\label{fig:ex_6}
\end{sidewaysfigure*}
\begin{sidewaysfigure*}
\centering
\scalebox{1}[1]{
\includegraphics[width=\textwidth]{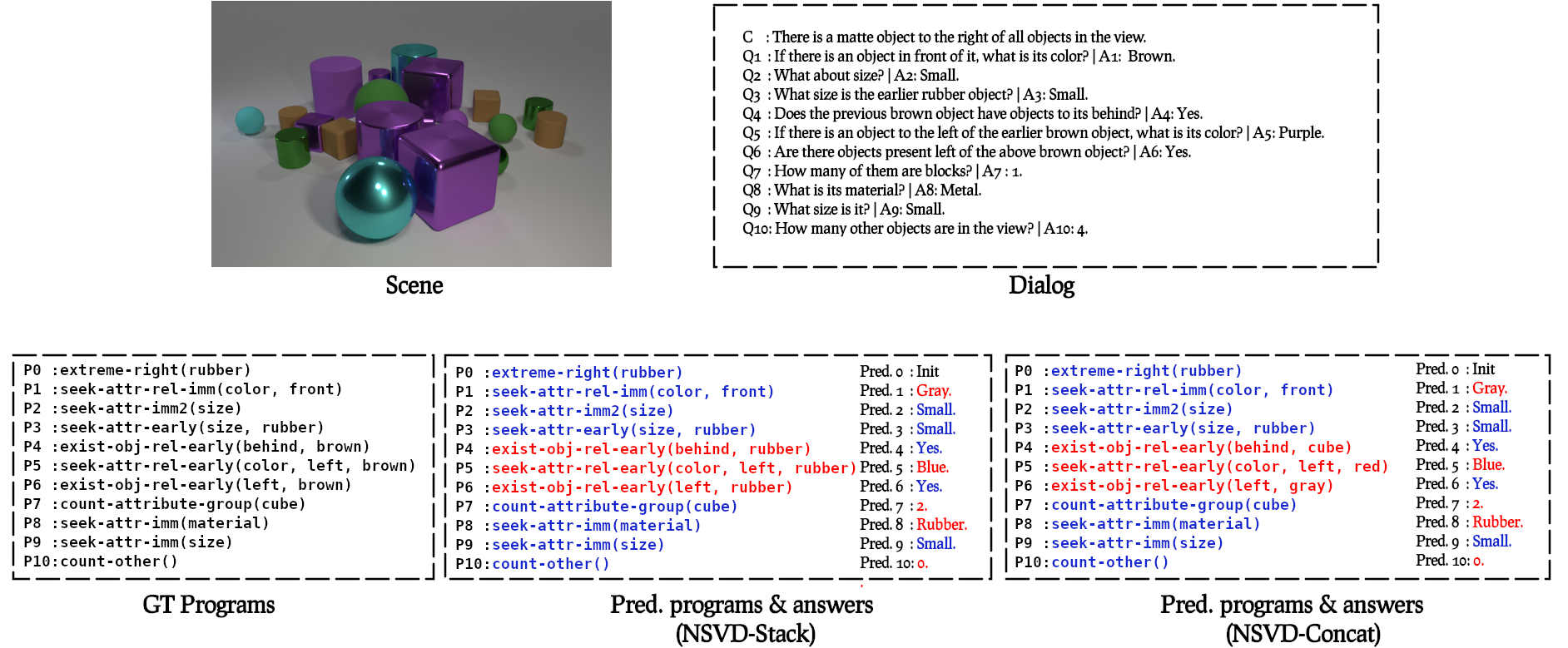}
}
\caption{Inference example of our models on a test instance with $20$ objects before fine-tuning. The blue and red colours indicate a match or a mismatch between the predicted program/answer and the ground truth, respectively.}
\label{fig:ex_7}
\end{sidewaysfigure*}

\begin{sidewaysfigure*}
\centering
\scalebox{1}[1]{
\includegraphics[width=\textwidth]{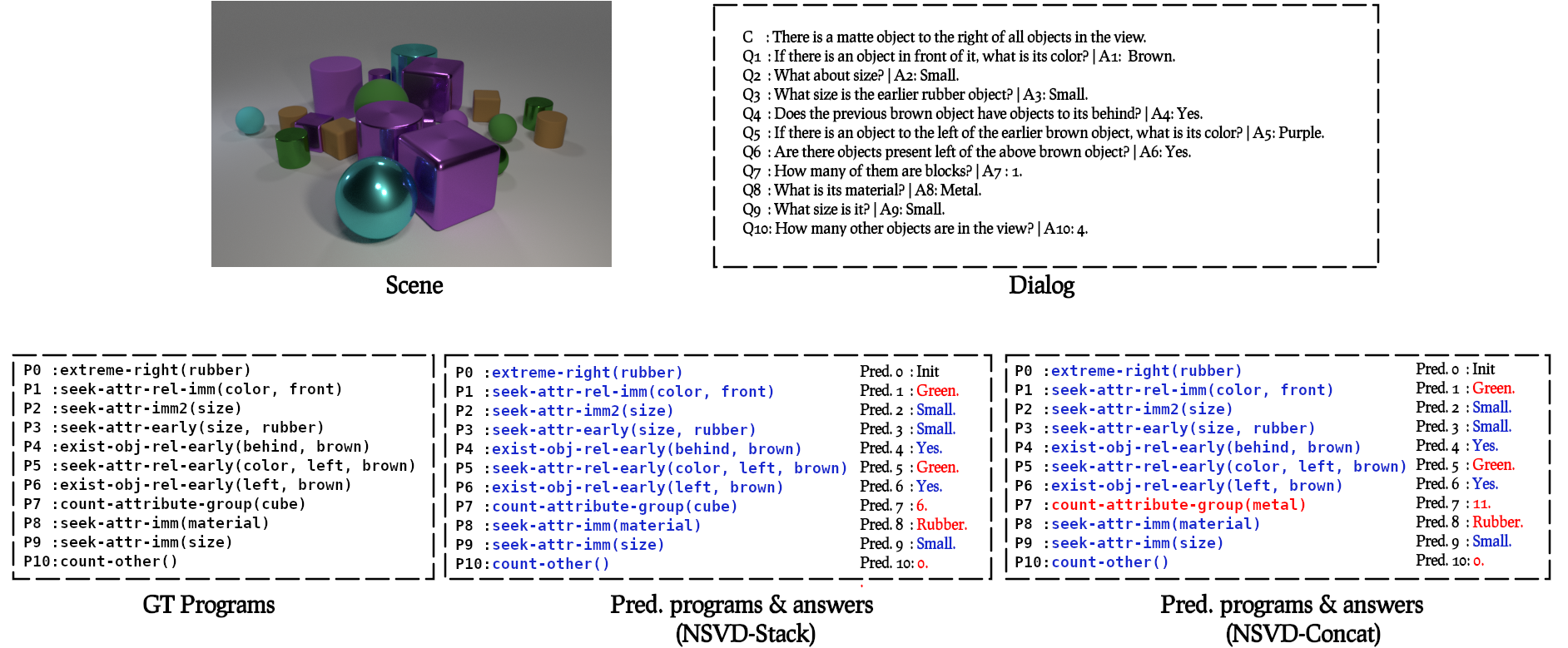}
}
\caption{Inference example of our models on a test instance with $20$ objects after fine-tuning. The blue and red colours indicate a match or a mismatch between the predicted program/answer and the ground truth, respectively.}
\label{fig:ex_8}
\end{sidewaysfigure*}
\begin{sidewaysfigure*}
\centering
\scalebox{1}[1]{
\includegraphics[width=\textwidth]{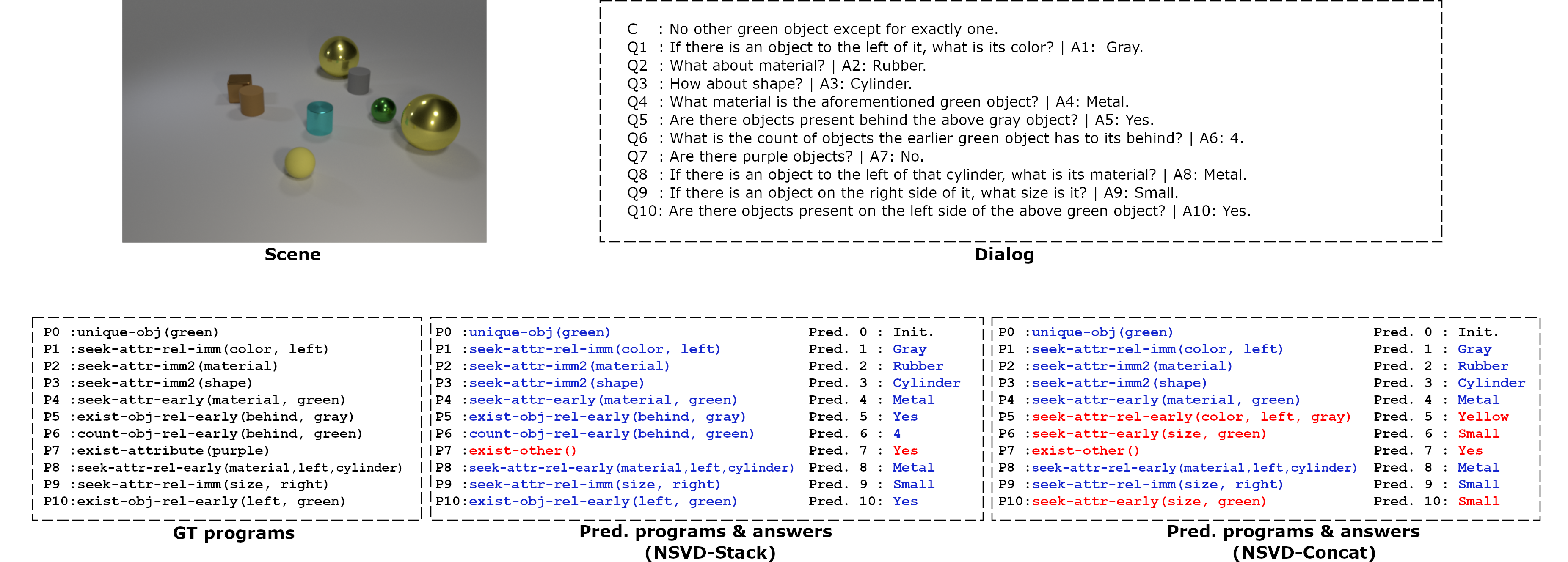}
}
\caption{Inference example of our models on a test instance of split BB before fine-tuning. While \textit{NSVD-stack} failed at answering only one question, \textit{NSVD-concat} failed at $4$ rounds. The blue and red colours indicate a match or a mismatch between the predicted program/answer and the ground truth, respectively.}
\label{fig:ex_9}
\end{sidewaysfigure*}

\begin{sidewaysfigure*}
\centering
\scalebox{1}[1]{
\includegraphics[width=\textwidth]{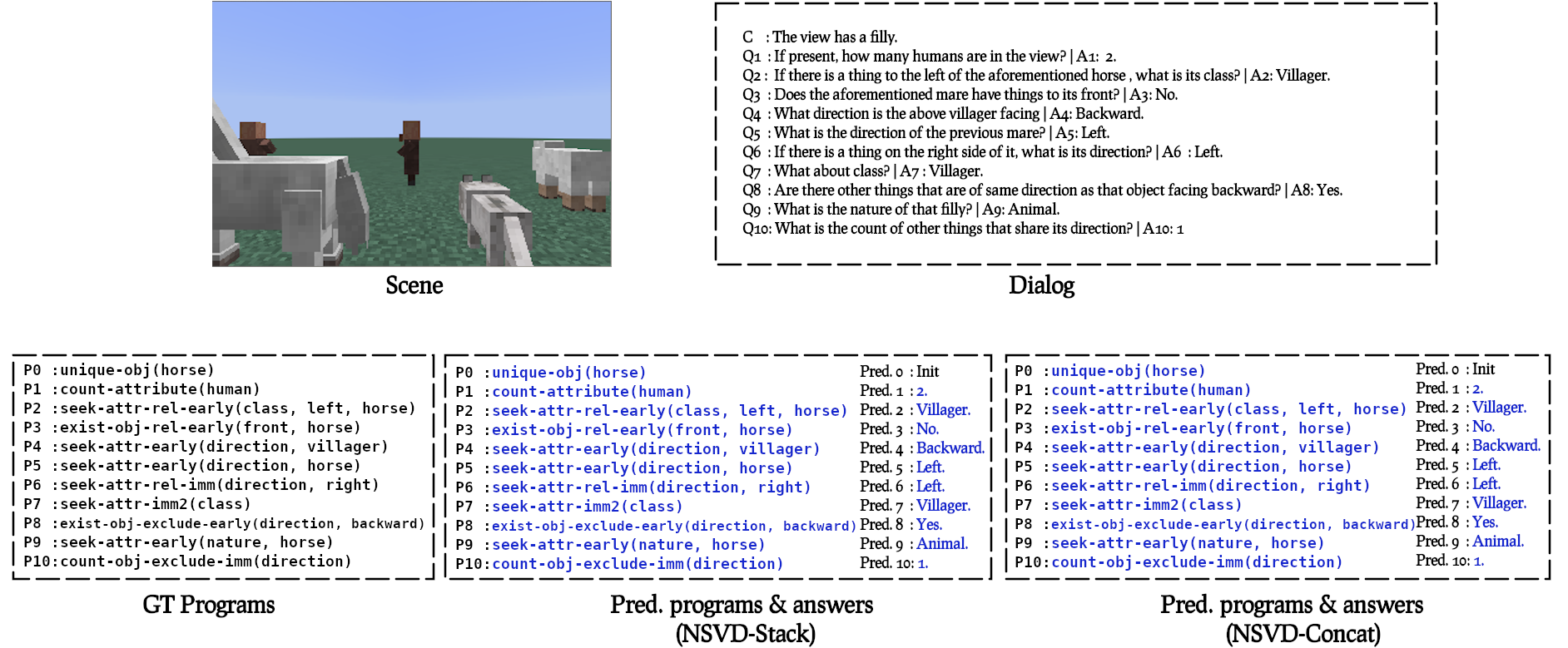}
}
\caption{Inference example of our models on a test instance of Minecraft-Dialog. The blue and red colours indicate a match or a mismatch between the predicted program/answer and the ground truth, respectively.}
\label{fig:ex_minecraft}
\end{sidewaysfigure*}

\end{document}